\documentclass[11pt]{article}
\usepackage[a4paper,hmargin=1in,vmargin=0.92in]{geometry}
\usepackage{amsmath,amssymb,amsthm,booktabs,graphicx}
\usepackage[hidelinks]{hyperref}
\input{glyphtounicode}
\hypersetup{pdftitle={Gaussian Self-Attention at Long Contexts: Sharp Rates, Score Rank, and Causal Masks},pdfauthor={Niles Yang}}
\newtheorem{theorem}{Theorem}
\newtheorem{lemma}[theorem]{Lemma}
\newcommand{\E}{\mathbb E}
\newcommand{\R}{\mathbb R}
\newcommand{\Pp}{\mathbb P}
\newcommand{\norm}[1]{\lVert #1\rVert}
\newcommand{\ind}{\mathbf 1}

\title{Gaussian Self-Attention at Long Contexts:\\
       \large Sharp Rates, Score Rank, and Causal Masks}
\author{Niles Yang}
\date{28 November 2026}
\begin{document}
\maketitle
\begin{abstract}
When each softmax row concentrates on a few keys, can averaging rows
still remove value noise? We study one attention head on a shared context of
$n$ i.i.d.\ Gaussian tokens in fixed dimension, under full and causal masks.
At fixed score norm $H$, the sharp worst-case exponent for the expected
absolute error of a pooled 1-Lipschitz statistic is $\min\{1/2,1/(1+H^2)\}$;
a two-coordinate head attains it for every $H$, answering the Gaussian case
of a sharpness question of Bohbot et al.\ (ICML 2026).

Now let the score matrix $H_nB_n$ grow with $n$, where $B_n$ has singular
values $1=\sigma_{1,n}\ge\cdots\ge\sigma_{d,n}$, and put
$a_n=\sqrt{2\log n}$. For i.i.d.\ centered unit-variance values independent
of the scores, the expected squared pooled mean is the probability that two
rows, sampled uniformly with replacement, draw the same key when each
samples from its own weights. For either mask and $H_n\ge ha_n$ with fixed
$h>0$, this risk is of order $\prod_{j=2}^d(1+a_n\sigma_{j,n})^{-1}$
uniformly in the spectrum, so for $d\ge2$ it vanishes if and only if
$a_n\sigma_{2,n}\to\infty$. When $H_n\to\infty$ and $H_n=o(a_n)$, it
acquires a further factor $\exp[-(2+o(1))\log n/H_n^2]$. In planar hardmax
attention, a weak second direction yields different full and causal
crossover curves, with rank-one endpoints $1/2$ and $1/4$.

On Qwen2.5-0.5B, at a matched byte budget, a 2/4-bit value cache improves
LAMBADA accuracy by 2.00 points over a 2-bit/BF16 allocation.
With bit counts and measured reconstruction errors held fixed, ranking
values by pooled rather than averaged squared attention changes accuracy by
only $-0.17$ points (95\% paired interval $[-0.60,0.27]$). With eight
16,384-token prefixes on an H100, the 2/4-bit cache cuts prefix KV storage
by 36.9\% and steady-state device memory by 7.4\%, at the cost of a 7.5\%
longer decode step than BF16 in the same implementation.
\end{abstract}

\section{Introduction}

An individual attention row and the mean output of a layer are built from
the same value vectors, but in different proportions. Suppose that ten
queries each select a single key. Their average carries the noise of ten
independent values if the selected keys are distinct, but the noise of only
one value if all ten queries select the same key. For centered independent
values of unit variance this distinction can be measured exactly:
conditional on the scores, the variance of the pooled output equals the
probability that two rows, chosen independently and uniformly, select the
same key when each row samples a key from its own softmax weights. We ask
how often such agreement occurs when all queries search the same Gaussian
context.

Our results give the sharp rate of this agreement. They cover score
strengths below the scale at which Gaussian softmax rows freeze, meaning
that a bounded number of keys begins to carry a positive fraction of a
row's weight, as well as strengths at and beyond that scale. Once the
strength grows with $n$, the rank of the score matrix matters, but only
through directions that are strong enough: a head can have two nonzero
score directions at every context length while the second one is still
too weak to make the pooled risk vanish. We identify the threshold at which
this happens and compute the limiting full and causal risks there. We also
track the prediction error when the values are the score coordinates
themselves. Finally, Section~\ref{sec:cache} tests
the resulting column weights in a trained decoder, where the quantization
residuals and the future queries are determined by the model rather than by
a Gaussian assumption.

\paragraph{Organization.} The rest of this section states the results for
a fixed score matrix. Section~\ref{sec:growing-results} lets the score
matrix grow as $A_n=H_nB$ and studies the transition at $H_n$ of order
$\sqrt{\log n}$; Section~\ref{sec:spectrum} further allows the singular
values and singular directions of the normalized score matrix to change
with $n$. Section~\ref{sec:sketch} explains why the rank of the score
matrix enters the pooled risk, Section~\ref{sec:cache} reports the
value-cache experiment, and Section~\ref{sec:related} discusses related
work and the scope of the results. All proofs are in the appendices.

\paragraph{Model.}
Let $X_1,\ldots,X_n$ be independent $\mathcal N(0,I_d)$ vectors. For fixed
matrices $A\in\R^{d\times d}$ and $V\in\R^{k\times d}$, set
\[
 F_n(x)=\frac{\sum_{j=1}^n e^{X_j^\top A x}V X_j}
                    {\sum_{j=1}^n e^{X_j^\top A x}},
 \qquad F(x)=V A x,\qquad H=\norm{A}_{\mathrm{op}} .
\]
Here $F_n(x)$ is the attention output for a query $x$, and $F$ is its
population limit; the formula for $F$ follows from Gaussian exponential
tilting, since $\E[e^{X^\top Ax}X]=e^{\norm{Ax}^2/2}Ax$ and
$\E e^{X^\top Ax}=e^{\norm{Ax}^2/2}$. In full self-attention the output at
position $i$ is $F_n(X_i)$, which includes the key and value of position $i$
itself. Under a causal mask the output is $F_i(X_i)$, which uses only the
keys and values at positions $1,\ldots,i$. Gaussian tokens with covariance
$\Sigma\succ0$ reduce to this model by whitening, which replaces $A$ and $V$
by $\Sigma^{1/2}A\Sigma^{1/2}$ and $V\Sigma^{1/2}$, so that $H$ coincides
with the attention horizon of \cite{bohbot}. The dimensions and the whitened
covariance are fixed throughout, and in this section $A$ and $V$ are fixed
as well. We write $Y_n=n^{-\gamma+o_{\mathbb P}(1)}$ if, for each fixed
$\eta>0$, the probability that $Y_n\in[n^{-\gamma-\eta},n^{-\gamma+\eta}]$
tends to one.

\begin{theorem}[Pooled Lipschitz statistics]\label{thm:lipschitz}
Fix $A$ and $V$. For a real $1$-Lipschitz function $h$ on $\R^k$, let
\[
 L_n(h)=\left|\frac1n\sum_{i=1}^n h(F_n(X_i))-\E h(F(X))\right|,
 \qquad X\sim\mathcal N(0,I_d).
\]
Then $\E L_n(h)\le n^{-\beta(H)+o(1)}$ uniformly in $h$, where
\[
 \beta(H)=\min\left\{\frac12,\frac{1}{1+H^2}\right\}.
\]
Conversely, for each $H\ge0$ the head on $\R^2$ with
$A=\operatorname{diag}(H,0)$ and $V=(0,1)$, together with $h(t)=|t|$,
satisfies $L_n(h)=n^{-\beta(H)+o_{\mathbb P}(1)}$ and
$\E L_n(h)=n^{-\beta(H)+o(1)}$.
\end{theorem}

\noindent The squared errors are governed by the exponents
\[
 \gamma_j(H)=\min\left\{1,\frac{\sqrt{1+4j/H^2}-1}{2}\right\}
 \quad(H>0),\qquad \gamma_j(0)=1,\qquad j=1,2 .
\]

\begin{theorem}[Expected error at a single token]\label{thm:row}
Fix $A$ and a nonzero $V$. Then
\[
 \E\norm{F_n(X_1)-F(X_1)}^2=n^{-\gamma_1(H)+o(1)} .
\]
The same exponent holds when the query is drawn independently of the
context.
\end{theorem}

The matching lower bounds use a particularly simple head, which we call the
\emph{marked head}. Each token consists of score coordinates together with
an independent scalar mark, and the head reads the mark as its value. Since
the population output of this head is zero, its outputs are themselves
errors.

\begin{theorem}[Typical and expected errors in the marked head]
\label{thm:marked}
Let the tokens be $(G_i,Z_i)$, where all $G_i\sim\mathcal N(0,I_d)$ and
$Z_i\sim\mathcal N(0,1)$ are independent. Query $i$ gives key $j$ the score
$G_j^\top A G_i$, and the value of key $j$ is $Z_j$. Write $Y_i$ for the
attention output at position $i$, and set $Q_n=n^{-1}\sum_i Y_i^2$ and
$P_n=(n^{-1}\sum_iY_i)^2$. Then $Q_n$ and $P_n$ obey the following power
laws, in probability ($Q_n=n^{-e+o_{\mathbb P}(1)}$) and in expectation
($\E Q_n=n^{-e+o(1)}$), and likewise for $P_n$:
\[
\begin{array}{lcc}
\toprule
\text{Squared error} & \text{Exponent in probability}
                     & \text{Exponent of expectation}\\
\midrule
Q_n\text{, individual tokens}&\min\{1,\,2/(1+2H^2)\}&\gamma_1(H)\\
P_n\text{, pooled mean}&\min\{1,\,2/(1+H^2)\}&\gamma_2(H)\\
\bottomrule
\end{array}
\]
\end{theorem}

\begin{figure}[t]
\centering
\includegraphics[width=\linewidth]{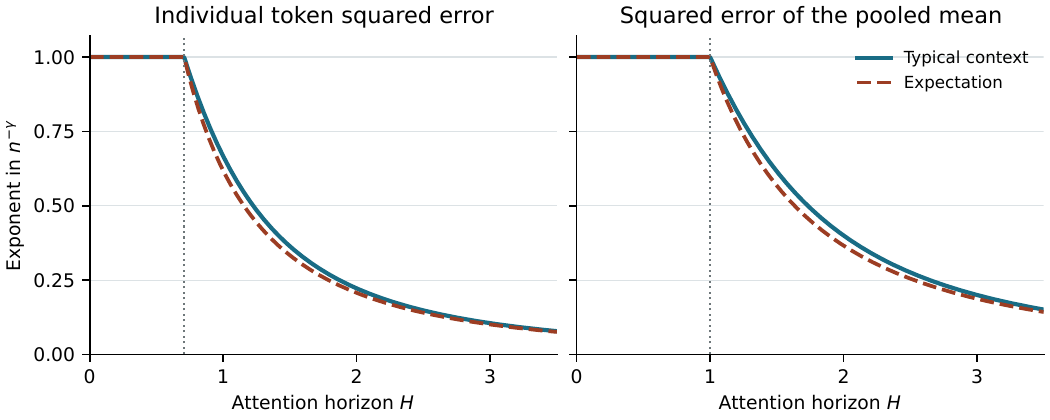}
\caption{Exponents $\gamma$ of the two squared errors in
Theorem~\ref{thm:marked}, in a typical context (solid) and in expectation
(dashed). At the dotted lines the power law leaves the parametric regime
$\gamma=1$. For Lipschitz pooling, $\beta(H)$ is half the typical exponent
of the squared pooled mean.}
\label{fig:phases}
\end{figure}

The thresholds are $H=1/\sqrt{2}$ for the tokenwise squared error and $H=1$
for the squared pooled mean. Above each threshold, the exponent of the
expectation is strictly smaller than the typical exponent
(Figure~\ref{fig:phases}): the expectation is then driven by rare events in
which an atypically large query meets an outlying key.
Theorems~\ref{thm:lipschitz} and~\ref{thm:row} answer the Gaussian case of
the sharpness question posed in Section~7 of \cite{bohbot}. Bohbot et al.\
bound each pooled Lipschitz observable with high probability (their
Theorem~5.1) and also consider the tokenwise error (their Appendix~D).
Theorem~\ref{thm:lipschitz} instead bounds $\sup_h\E L_n(h)$ for a fixed
head and shows that, for every $H$, the bound is attained by a specific
head and a fixed observable.

\begin{theorem}[Causal self-attention]\label{thm:causal}
For the causal outputs $F_i(X_i)$, Theorem~\ref{thm:lipschitz} holds with
the same exponent and the same extremal head. In the setting of
Theorem~\ref{thm:row}, both the error at the last position and the average
of the individual expected squared errors over all $n$ positions have
exponent $\gamma_1(H)$. All four exponents of the marked head in
Theorem~\ref{thm:marked} are unchanged.
\end{theorem}

\begin{theorem}[The boundary of the parametric regime]\label{thm:critical}
In the setting of Theorem~\ref{thm:row}, let $R_n$ denote the expected
individual squared error, for either an independent query or the query from
the context. If $H<1/\sqrt2$, then
\[
 \lim_{n\to\infty}nR_n=
 \E_{X\sim\mathcal N(0,I_d)}
       e^{\norm{AX}^2}\big(\norm{V}_{\mathrm F}^2+\norm{VAX}^2\big).
\]
At $H=1/\sqrt2$, let $\mathcal E$ be the eigenspace of $A^\top A$ for its
largest eigenvalue, let $\mu=\dim\mathcal E$, and let
$\iota=\ind\{VA|_{\mathcal E}\ne0\}$. Then
$R_n=\Theta(n^{-1}(\log n)^{\mu/2+\iota})$. For causal attention, the
average risk over all positions is of order $n^{-1}\log n$ in the
subcritical case and of order $n^{-1}(\log n)^{\mu/2+\iota+1}$ at the
threshold.
\end{theorem}

\section{When the score strength grows with context length}
\label{sec:growing-results}

The fixed-$H$ exponents of Section~1 depend on the score matrix only through
its operator norm. Its rank appears once the score strength grows with the
context length. Throughout this section we write $a_n=\sqrt{2\log n}$, fix a
score matrix $B$ with $\|B\|_{\mathrm{op}}=1$ and rank $r\ge1$, and use
$A_n=H_nB$ with a deterministic $H_n\ge0$; in a causal forward pass all
prefixes share the same $A_n$. We work with the marked head of
Theorem~\ref{thm:marked}, whose tokens consist of score coordinates $G_i$
and an independent standard Gaussian scalar value $Z_i$, so that the
population output is zero. Besides $Q_n=n^{-1}\sum_iY_i^2$ and
$P_n=(n^{-1}\sum_iY_i)^2$, we consider $M_n=n^{-1}\sum_i|Y_i|$. A row is
\emph{concentrated}, or \emph{frozen}, if for large $n$ a bounded number of
keys carries a positive fraction of its weight. We write
$Y_n=\Theta_{\mathbb P}(x_n)$ when both $Y_n/x_n$ and $x_n/Y_n$ are tight,
and constants may depend on $B$ and on the fixed dimension.

\begin{theorem}[Growing strength and the rank of the score map]\label{thm:growing}
The following conclusions hold for both full and causal attention.
\begin{enumerate}
\item Suppose that $H_n\to\infty$ and $H_n=o(a_n)$, and set
$T_n=\log n/H_n^2$. Then
\[
 \frac{1}{T_n}\bigl(\log\E M_n,\ \log\E Q_n,\
                \log(a_n^{r-1}\E P_n)\bigr)\ \longrightarrow\ (-1,-1,-2).
\]
The same convergence holds in probability when $M_n,Q_n,P_n$ replace their
expectations.
\item If $\liminf H_n/a_n>0$, then
\[
  \E P_n=\Theta(a_n^{-(r-1)}),\qquad
  P_n=\Theta_{\mathbb P}(a_n^{-(r-1)}),\qquad Q_n=\Theta_{\mathbb P}(1).
\]
If moreover $r=1$, then $\liminf\operatorname{Var}(Q_n)>0$.
\item If $H_n/a_n\to c\in[0,\infty]$, then
\[
 \E Q_n\longrightarrow q_c(B),\qquad
 q_c(B)=\E\left(1-\frac{1}{c\|BG\|}\right)_+
 \quad(0<c<\infty),\quad G\sim\mathcal N(0,I_d),
\]
with $q_0(B)=0$ and $q_\infty(B)=1$. If $r\ge2$, then also
$Q_n\to q_c(B)$ in $L^2$.
\end{enumerate}
\end{theorem}

For example, $H_n^2=\log n/(\lambda\log\log n)$ with $\lambda>0$ gives
$Q_n=(\log n)^{-\lambda+o_{\mathbb P}(1)}$ and
$P_n=(\log n)^{-2\lambda-(r-1)/2+o_{\mathbb P}(1)}$: each additional
direction of the score map improves the logarithmic exponent of the squared
pooled mean by $1/2$, an effect that the fixed-$H$ power laws cannot see.
The theorem also yields consistency criteria that apply to strength
sequences without a limit.
\begin{itemize}
\item For every rank, $\E Q_n\to0$ if and only if $H_n=o(a_n)$.
\item For $r=1$, $\E P_n\to0$ if and only if $H_n=o(a_n)$. For $r\ge2$, by
contrast, $\E P_n\to0$ for every sequence $H_n$, with the uniform upper
bound $\E P_n=O_B((1+\log n)^{-(r-1)/2})$.
\end{itemize}
The criteria for $\E Q_n$ and for rank one follow from
Theorem~\ref{thm:growing}, since $P_n\le Q_n$ and any subsequence along
which $H_n/a_n$ stays bounded away from zero falls under part~2; the
uniform bound for $r\ge2$ is the two-query estimate of
Section~\ref{sec:angle}. Because the outputs have
uniformly bounded Gaussian fourth moments, the same criteria hold for
convergence in probability. For rank one, the two orders of limits
therefore differ: at any fixed strength the pooled risk tends to
zero, whereas passing to hardmax at each length first gives the positive
limits of Theorem~\ref{thm:records}.

\begin{figure}[t]
\centering
\includegraphics[width=\linewidth]{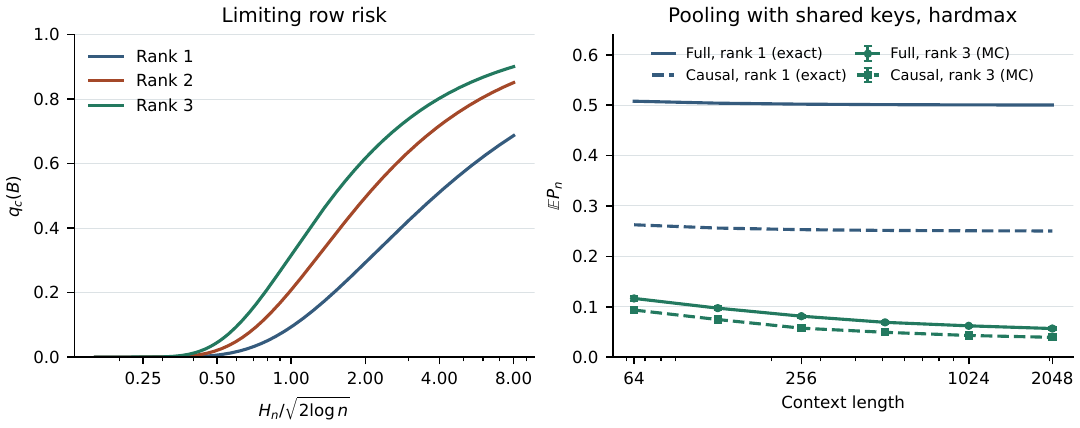}
\caption{Left: the limiting row risk $q_c(B)$ of
Theorem~\ref{thm:growing} when all nonzero singular values of $B$ equal one.
Right: exact rank-one hardmax risks, and rank-three simulations in which all
rows share the same isotropic key list. Each token has independent key,
query and value coordinates. Each point is an average over 80 contexts,
with the values integrated out analytically conditional on the scores. Bars
show 1.96 Monte Carlo standard errors.}
\label{fig:growing}
\end{figure}

\begin{theorem}[Outputs with a general value map]\label{thm:clipping}
Let the tokens $X_i$ be as in Theorem~\ref{thm:row}, let $V$ be any fixed
linear value map, let $O_i$ be the full or causal output at position $i$,
and let $\pi(y)=y/\max(1,\|y\|)$ be the projection onto the closed unit
ball. For any deterministic sequence $H_n\ge0$, put $c_n=H_n/a_n$. Then
\[
 \frac1n\sum_i\left\|\frac{O_i}{a_n}-V\pi(c_nBX_i)\right\|^2
       \xrightarrow{L^1}0.
\]
If $c_n\to c<\infty$, the normalized error with respect to the population
output satisfies
\[
 \frac1{n a_n^2}\sum_i\|O_i-H_nVBX_i\|^2
       \xrightarrow{L^1}\E\|V(\pi(cBX)-cBX)\|^2,
\]
and for $c>0$ this limit is positive if and only if $VB\ne0$. If
$c_n\to\infty$, then
$\frac1{nH_n^2}\sum_i\|O_i-H_nVBX_i\|^2\to\|VB\|_{\mathrm F}^2$ in $L^1$.
\end{theorem}

The marked head isolates the value noise along coordinates orthogonal to
the score space. Theorem~\ref{thm:clipping} also captures the systematic
error along the score directions: once the query norm is large enough, the
empirical softmax barycenter reaches the boundary of the scaled Gaussian
sample, so the weighted key average saturates at norm about $a_n$ instead
of growing linearly with the query.

\begin{theorem}[Rank-one hardmax limit]\label{thm:records}
Consider the marked head with $r=1$ and $H_n/a_n\to\infty$. Under full
attention, $\overline Y=n^{-1}\sum_iY_i$ converges in law to
$\mathcal N(0,1/2)$. Under causal attention, it converges in law to a
centered Gaussian variance mixture with random variance $W=(S_++S_-)/4$,
where $S_+,S_-$ are independent copies of
\[
 S=\sum_{j\ge1}(R_{j-1}-R_j)^2,\qquad R_0=1,\quad
 R_j=R_{j-1}U_j,\quad U_j\ \text{i.i.d. uniform on }[0,1].
\]
In particular, the limits of $\E P_n$ are $1/2$ (full) and $1/4$ (causal),
and $\operatorname{Var}(W)=1/192$. The same limits hold under exact hardmax
selection at every $n$. Under full attention the tokenwise risk $Q_n$
converges in law to an exponential variable with mean one; under a causal
mask its limiting variance is $1/2$.
\end{theorem}

The two masks share all fixed-$H$ exponents. At high score strength,
however, the extreme key of a causal prefix keeps being replaced as the
prefix grows, and these successive records change the limiting
distribution.

\begin{theorem}[The key projection used as the value]\label{thm:aligned}
Write $B=uv^\top$ with unit vectors $u,v$, and let the value of key $j$ be
its own score projection $S_j=u^\top X_j$. Let $\overline O$ be the pooled
scalar output. For exact hardmax, and for softmax under the sufficient
condition $H_n/(a_n\log n\log\log n)\to\infty$,
\[
 \lim_{n\to\infty}(\log n)\E\overline O^{\,2}=
 \begin{cases}\pi^2/24&\text{full},\\1/4&\text{causal}.\end{cases}
\]
Under full attention, $a_n\overline O$ converges in law to half the
difference of two independent standard Gumbel variables. Under causal
attention, it converges in law to
$\tfrac12\int_0^1 (\mathcal M_+(s)-\mathcal M_-(s))\,ds$, where
$\mathcal M_+$ and $\mathcal M_-$ are independent copies of the running
maximum $\mathcal M(s)=\max\{x:(t,x)\in\Pi,\ t\le s\}$ of a Poisson point
process $\Pi$ on $(0,1]\times\R$ with intensity $dt\,e^{-x}dx$.
\end{theorem}

For $u=v$, the full-hardmax constant recovers Proposition~5.3 of
\cite{bohbot}.
Here the causal-to-full ratio of the asymptotic risks is $6/\pi^2$, whereas
for independent values Theorem~\ref{thm:records} gives the ratio $1/2$; the
two choices of value also lead to different orders in $n$. The softmax
condition in Theorem~\ref{thm:aligned} suffices for this finer logarithmic
scale, but its sharp threshold remains open. The proofs are given in
Section~\ref{sec:growing-proof}.

The expectation formulas for $P_n$ and $Q_n$ in both strength regimes hold
for any i.i.d.\ centered marks of unit variance that are independent of the
scores, because the conditional second moments depend only on the
covariance of the marks. Probability statements, by contrast, depend on the
law of the marks. For example, take symmetric sign marks and full rank-one
hardmax. With probability $1/2$ the two extreme marks agree, and then
$P_n=1$. On the complementary event $P_n=(2N_+/n-1)^2$, where $N_+$ is the
number of queries with positive sign; since the two marks are independent
of the score tokens, $nP_n$ converges in law to $\chi^2_1$ on that event.
The same limiting expected risk $1/2$ thus arises from two different
scales.

\subsection{When a score direction becomes weak}\label{sec:spectrum}

The constants in Theorem~\ref{thm:growing} may depend on the smallest
nonzero singular value of $B$. To see how a weak score direction affects the
pooled risk, consider a deterministic sequence $B_n$ in fixed dimension $d$
with singular values $1=\sigma_{1,n}\ge\cdots\ge\sigma_{d,n}\ge0$; the
singular vectors may also vary with $n$. Define
\[
      N_n=\prod_{k=2}^d(1+a_n\sigma_{k,n}),\qquad
      a_n=\sqrt{2\log n}.
\]
Full and causal attention use $A_n=H_nB_n$, together with the independent
scalar Gaussian values of the marked head (Theorem~\ref{thm:marked}).

\begin{theorem}[A bound uniform in the singular values]\label{thm:spectrum}
For either mask, $\E P_n\le C_d/N_n$ for every strength $H_n$. Moreover,
for each $h>0$, all sufficiently large $n$ and every $H_n\ge h a_n$,
\[
       \frac{c_{d,h}}{N_n}\le \E P_n\le\frac{C_d}{N_n},
       \qquad P_n=\Theta_{\mathbb P}(N_n^{-1}).
\]
These bounds also hold for exact hardmax. The constants and the tightness
bounds are uniform over the singular values and the singular vectors. For
$d\ge2$, under $H_n\ge h a_n$ or exact hardmax,
\[
 \operatorname{Var}(Q_n)\to0
       \quad\Longleftrightarrow\quad a_n\sigma_{2,n}\to\infty .
\]
Finally, if $H_n\to\infty$ and $H_n=o(a_n)$, put $T_n=\log n/H_n^2$; then
\[
 \frac{\log(N_n\,\E P_n)}{T_n}\longrightarrow -2,\qquad
 \frac{\log(N_n P_n)}{T_n}\xrightarrow{\mathbb P}-2 .
\]
\end{theorem}

Thus, for $d\ge2$ and $H_n\ge h a_n$, the squared pooled mean vanishes in
expectation if and only if $a_n\sigma_{2,n}\to\infty$, and each further
singular direction contributes its own factor. For example, singular values
$(1,a_n^{-1/2},a_n^{-2})$ give $\E P_n=\Theta((\log n)^{-1/4})$, although
the rank is three at every $n$. Under the same condition on $H_n$, or for
hardmax, a planar head with singular values $(1,\varepsilon)$ has pooled
risk of order $(1+\varepsilon\sqrt{2\log n})^{-1}$, with constants that do
not depend on $\varepsilon$; its transition therefore occurs on the scale
$\log n\asymp\varepsilon^{-2}$.

Inside this transition the limiting risk also depends on the mask. For
$s\ge0$, $0<q\le1$ and $Z\sim\mathcal N(0,1)$, set
\begin{equation}\label{eq:prefix-concurrence}
 K_s(q)=\E_Z\!\left[
 \frac{1}{\Phi(Z)+e^{s^2/2-sZ}\{\Phi(s-Z)+q^{-1}-1\}}\right].
\end{equation}
As shown in Section~\ref{sec:prefix-proof}, $K_s(q)$ is the limiting
probability that two hardmax rows select the same key when the angle
$\theta$ between their query directions satisfies $z_n\sin\theta\to s$,
where $z_n=\Phi^{-1}(1-1/n)$, and $q$ is the length of the shorter prefix
divided by that of the longer one. At $q=1$ this is the Brown--Resnick
extremal concurrence probability of Dombry et al.~\cite[Example~6]{concurrence};
the term $q^{-1}-1$ accounts for the keys that only the later query can
see.

\begin{theorem}[Planar crossover]\label{thm:crossover}
Let $B_n=\operatorname{diag}(1,\varepsilon_n)$ with $0\le\varepsilon_n\le1$
and $a_n\varepsilon_n\to\lambda\in[0,\infty)$. Under exact hardmax, or under
softmax with $H_n/a_n\to\infty$, the limits of $\E P_n$ for full and causal
attention are, respectively,
\[
 \mathcal F(\lambda)=\tfrac12\E_D K_{\lambda |D|}(1),
 \qquad
 \mathcal C(\lambda)=\tfrac12\E_D\int_0^1 K_{\lambda |D|}(q)\,dq,
\]
where $D$ is a centered Cauchy variable with scale two. Both functions are
continuous, $\mathcal F(0)=1/2$, $\mathcal C(0)=1/4$, and
$0<\mathcal C(\lambda)<\mathcal F(\lambda)$ for every finite $\lambda$. Both
are of order $(1+\lambda)^{-1}$ as $\lambda\to\infty$.
\end{theorem}

\begin{figure}[!b]
\centering
\includegraphics[width=.91\linewidth]{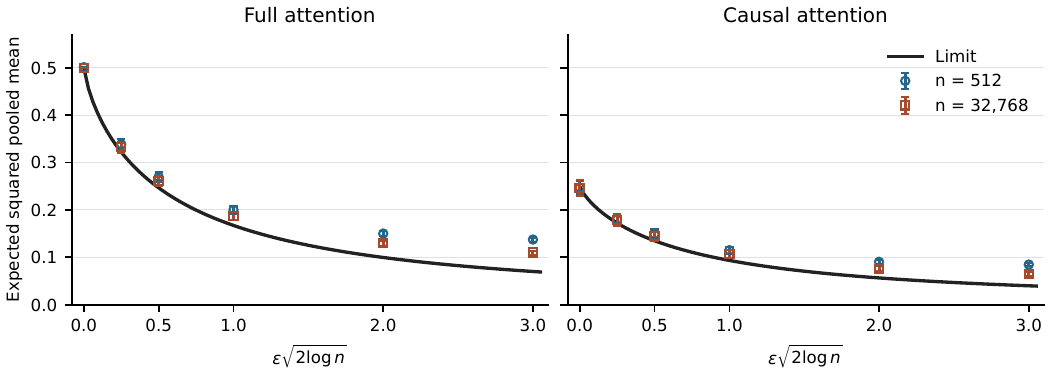}
\caption{The limiting risks of Theorem~\ref{thm:crossover} (curves) and
finite planar hardmax self-attention (points). Each point averages 96
independent contexts, with the values integrated out exactly; error bars
are 1.96 Monte Carlo standard errors. Keys and queries are the same
tokens.}
\label{fig:spectrum}
\end{figure}

At $\lambda=1$ the full and causal limits are $0.1672$ and $0.0934$, while
at $n=32{,}768$ the corresponding finite-context estimates are $0.1874$ and
$0.1070$ (standard errors $0.0043$ and $0.0041$).
Figure~\ref{fig:spectrum} shows this finite-length discrepancy for six
values of $\lambda$. The proofs are in Sections~\ref{sec:spectrum-proof}
and~\ref{sec:prefix-proof}.

\section{Why the score rank enters the pooled risk}\label{sec:sketch}

This section outlines the proof of Theorem~\ref{thm:growing}; the details
are in Section~\ref{sec:growing-proof}. Let $w_{ij}$ be the attention weight
of query $i$ on key $j$, and write $p_j=n^{-1}\sum_i w_{ij}$ for the mean
weight assigned to key $j$. For independent values of unit variance, the
pooled risk conditional on the scores is exactly
$\sum_j p_j^2=n^{-2}\sum_{i,k}\langle w_i,w_k\rangle$, so the task is to
estimate the overlap $\langle w_i,w_k\rangle$ of two rows.

\paragraph{Upper bound.} Take two different queries and remove their own
indices from both key lists; this changes their expected overlap only by
a polynomially small amount, uniformly in $H_n$. Given the two queries, the
projected scores of each common key form a Gaussian pair whose correlation
is the inner product of the two query directions. If the absolute value of
this correlation is $s$, Gaussian hypercontractivity bounds the expected
overlap by $m^{-(1-s)/(1+s)}$, where $m$ is the length of the shorter
remaining list. For late queries $m$ is comparable to $n$, so this bound is
appreciable only when the two directions lie within an angle of order
$(\log n)^{-1/2}$ of each other, up to sign. Integrating over directions in
the $r$-dimensional score space gives the factor $(\log n)^{-(r-1)/2}$.

\paragraph{Lower bound.} The reverse inequality counts the keys whose
projection onto the score space has norm at least $z_n-K/z_n$, where
$z_n=\Phi^{-1}(1-1/n)$ and $K$ is a fixed constant. There are only
$O_{\mathbb P}(a_n^{r-1})$ such keys. For $K$ large enough, a query with
score norm $H_n\|BG_i\|>(1+\delta)a_n$ puts a fixed fraction of its mass on
these keys with probability arbitrarily close to one. If such queries make
up a fraction $p_*$ of the context, the Cauchy--Schwarz inequality gives a
pooled risk of order at least $p_*^2/a_n^{r-1}$. Below the freezing scale,
each such query has Gaussian tail probability
$\exp[-(1+\delta)^2T_n+o(T_n)]$, and the matching upper bound uses the same
tail for each of two independent queries. This explains why the coefficient
of $T_n$ doubles for pooling, and why the score rank still matters when
$T_n$ grows very slowly.

Figure~\ref{fig:growing} shows the size of this effect in finite contexts
and compares the two masks. At $n=2048$ under rank-three hardmax, the mean
squared pooled outputs are $0.0568$ (full) and $0.0394$ (causal), with Monte
Carlo standard errors $0.0011$ and $0.0009$. With independent query and key
coordinates, the rank-one values at this length are already within $0.001$
of their limits $1/2$ and $1/4$. The simulation only illustrates the
finite-context behavior; the exponents in Theorem~\ref{thm:growing} follow
from the two bounds above.

Table~\ref{tab:soft} reports finite-temperature simulations at $H_n=a_n$
with $B=I_r$, in which each query is also one of the keys and each value is
an independent scalar mark. At $n=1024$ both masks retain more individual
value noise than the limiting formula predicts, and the gap is larger for
the causal mask, whose rows include many short prefixes.

\begin{table}[ht]
\centering
\small
\begin{tabular}{cccccc}
\toprule
& \multicolumn{2}{c}{Individual risk $\E Q_n$} &
\multicolumn{2}{c}{Pooled risk $\E P_n$} & Row limit\\
$r$ & Full & Causal & Full & Causal & $q_1(I_r)$ \\
\midrule
1 & .162 (.008) & .203 (.007) & .0443 (.0026) & .0344 (.0021) & .0940\\
3 & .428 (.006) & .476 (.005) & .0362 (.0010) & .0274 (.0009) & .3173\\
\bottomrule
\end{tabular}
\caption{Finite-temperature self-attention with $n=1024$ and
$H_n=\sqrt{2\log n}$. Each entry averages 64 independent Gaussian contexts,
with one Monte Carlo standard error in parentheses. The values are
integrated out analytically, conditional on the scores. The last column is
the limit $q_1(I_r)$ of $\E Q_n$ from Theorem~\ref{thm:growing}.}
\label{tab:soft}
\end{table}

\section{A fixed-budget test on a trained value cache}\label{sec:cache}

In the Gaussian model the pooled risk is governed by the column masses,
that is, by how much attention each key receives on average. This suggests
a rule for deciding which values to store precisely. Consider one KV head
shared by $G$ query heads. For $T$ calibration queries, let $a_{htj}$ be the
causal attention that head $h$ pays to prefix position $j$ at query $t$,
with zero weight on positions not yet seen. Set
\[
 p_{hj}=\frac1T\sum_t a_{htj},\qquad
 g_j=\frac1G\sum_h p_{hj}^2,\qquad
 r_j=\frac1{GT}\sum_{h,t}a_{htj}^2,\qquad m_j=\frac1G\sum_h p_{hj}.
\]
Suppose that the quantization errors $e_j$ are centered and, conditional on
the weights, independent across positions, with
$\Delta_j=\E[\norm{e_j}^2\mid a]$. Expanding the square then gives
\[
 \frac1G\sum_h \E\!\left[\left\|\sum_j p_{hj}e_j\right\|^2
             \,\middle|\, a\right]=\sum_j g_j\Delta_j .
\]
If instead the individual query errors are squared before averaging, the
same expansion gives $\sum_j r_j\Delta_j$. The two coefficients differ by
$r_j-g_j=G^{-1}\sum_h T^{-1}\sum_t(a_{htj}-p_{hj})^2$, the temporal variance
of the attention paid to position $j$. Hence, at equal mean attention, the
row criterion gives more weight to positions that receive isolated spikes
of attention. These identities hold for any fixed weights; the Gaussian
theorems determine how the pooled coefficient sum scales within their own
model.

\paragraph{Fixed prefix storage.}
We use Qwen2.5-0.5B~\cite{qwen} in BF16, which has 24 layers, two KV heads
per layer, seven query heads per KV head and value dimension $d=64$. For a
prefix of length $P$, all scores use only its last $T=\min(64,P)$ queries.
Let $E_{b,j}=\norm{v_j-\widehat v_{b,j}}^2$ be the error of a $b$-bit affine
quantizer with round-to-nearest and BF16 reconstruction, and write
$D_j=E_{2,j}-E_{4,j}$ and $S_j=\sqrt{E_{2,j}/d}-\sqrt{E_{4,j}/d}$. For a
fixed number of 4-bit upgrades, an exchange argument shows that promoting
the values with the largest $g_jD_j$ minimizes $\sum_j g_jE_{b_j,j}$. This
objective is only a proxy: for deterministic residuals the squared output
also contains cross terms, the ranking uses past queries as a stand-in for
future ones, and the trained model changes its later attention weights.

Let $u=\lfloor P/8\rfloor$. The 2/BF16 baseline keeps $u$ values in BF16,
chosen by $g_j$ times their estimated stochastic-rounding variance. The 2/4
policy assigns 4 bits to $\lfloor27u/4\rfloor$ values. The 2/4/8 policy
assigns 4 bits to $\lfloor15u/4\rfloor$ values and 8 bits to $u$ values: it
first selects all promotions by $g_jD_j$ and then selects the 8-bit subset
by $g_j(E_{4,j}-E_{8,j})$. In every policy the remaining values receive 2
bits, and all prefix keys as well as all new keys and values stay in BF16.
At $P=512$ each of the three mixed policies costs 82,688 prefix KV bytes per
layer and KV head, including the two BF16 affine parameters of each
quantized vector. At other lengths the 2/4 and 2/4/8 schedules have
identical costs, at most 12 bytes per layer and KV head below the cost of
2/BF16.

\begin{table}[htbp]
\centering\small
\begin{tabular}{@{}llrrrr@{}}
\toprule
Value bits & Allocation score & KV (KiB) & Words correct & PPL 512 & PPL 4096\\
\midrule
BF16 & all values & 128.00 & 2,682 & 11.5091 & 10.9365\\
2/BF16 & protected subset & 80.75 & 2,571 & 11.6195 & 10.9998\\
2/4/8 & two pooled upgrades & 80.75 & 2,634 & 11.5313 & 10.9420\\
2/4 & attention mass $m_j$ & 80.75 & 2,647 & 11.5515 & 10.9691\\
2/4 & pooled $g_jD_j$ & 80.75 & 2,674 & 11.5501 & 10.9726\\
2/4 & row $r_jD_j$ & 80.75 & 2,683 & 11.5561 & 10.9712\\
2/4 & mass $m_jD_j$ & 80.75 & 2,665 & 11.5556 & 10.9794\\
2/4 & mass $m_jS_j$ & 80.75 & 2,665 & 11.5561 & 10.9805\\
4 & all values & 82.00 & 2,697 & 11.5402 & 10.9590\\
\bottomrule
\end{tabular}
\caption{Prefix K and V storage per layer and KV head at 512 tokens, number
of LAMBADA final words predicted exactly (out of 5,153), and WikiText-2 test
perplexity at two prefix lengths. All 2/4 rows share the same quantizer and
the same bit counts and differ only in the allocation score.}
\label{tab:cache-quality}
\end{table}

\paragraph{Task and score controls.}
On English LAMBADA~\cite{lambada} we hold back the last context token,
compress the preceding 46--221 tokens, and then decode greedily from the
packed cache; a prediction counts as correct only if every subword of the
final word matches. On WikiText-2~\cite{wiki} we score 128 subsequent
one-step predictions per article. Here the packed values are restored to
their original order, so that every condition uses the same BF16 attention
implementation. At $P=512$ and $P=4096$, respectively, 58 and 26 test
articles are long enough to qualify. We chose the 2/4 schedule after
inspecting the uniform 4-bit results, and afterwards also evaluated the
untouched validation partition (60 and 23 articles). The score controls
were added later, during the comparison with prior work.

The pooled 2/4 policy answers 103 more words correctly than 2/BF16, a
difference of $+2.00$ percentage points (95\% paired bootstrap interval
$[1.36,2.64]$). This contrast, however, changes the precision schedule as
well as the ranking rule. With the schedule held fixed, the pooled score
$g_jD_j$ gains 27 words over plain attention mass $m_j$, or $+0.52$ points
$[0.12,0.95]$. When the measured distortion gain is held fixed as well,
the pooled score answers nine fewer words than the row score $r_jD_j$, or
$-0.17$ points
$[-0.60,0.27]$, and the sign of their difference in test perplexity changes
with $P$ (Table~\ref{tab:cache-quality}). The 2/4 policy trails uniform
4-bit by 23 words, $-0.45$ points $[-0.99,0.12]$, while uniform 4-bit uses
1.55\% more prefix KV bytes than 2/4 at $P=512$. All intervals use 10,000
paired resamples by example or by article, without adjustment for
multiplicity. These comparisons do not establish a task-level advantage
that is specific to temporal pooling.

\begin{table}[htbp]
\centering\small
\begin{tabular}{@{}rrrrrrr@{}}
\toprule
 & & \multicolumn{3}{c}{ms per batch decode step} & \multicolumn{2}{c}{Device MiB}\\
\cmidrule(lr){3-5}\cmidrule(l){6-7}
Batch & Prefix & Flash BF16 & Split BF16 & Split 2/4 & Split BF16 & Split 2/4\\
\midrule
1 & 512    & 2.710 & 2.475 & 2.493 & 1,837 & 1,843\\
1 & 4,096  & 2.839 & 2.508 & 2.524 & 1,913 & 1,879\\
8 & 4,096  & 3.298 & 2.851 & 2.900 & 2,463 & 2,329\\
1 & 16,384 & 3.101 & 2.591 & 2.605 & 2,139 & 2,143\\
8 & 16,384 & 3.937 & 3.324 & 3.572 & 4,307 & 3,989\\
\bottomrule
\end{tabular}
\caption{Full-model greedy decoding on an H100 80~GB. The Flash column uses
FlashAttention 2.7.4; the two split columns use the same tiled attention
with softmax merging. Device memory is total minus free memory during steady
decoding on an exclusive GPU. Each step emits one token per sequence.}
\label{tab:cache-speed}
\end{table}

\paragraph{Memory and time.}
We pack four 2-bit codes per byte. Within each KV head, the post-rotary
prefix keys are reordered together with their values so that values of the
same bit width are stored contiguously; because every later query sees the
complete prefix, this common permutation leaves the attention output
unchanged. During decoding the kernel unpacks the values on the fly, and new
tokens are appended to a preallocated BF16 tail of 32 positions.
Table~\ref{tab:cache-speed} uses contiguous cross-layer buffers and fixed
synthetic prompts. Each process performs 15 timed CUDA Graph runs of 32
decoding steps after warm-up, including the output projection and greedy
selection. For the shapes $1\times4096$ and $8\times16384$ we report the
median of three medians from fresh processes; the other shapes use the
median of a single process.

At $8\times16384$, relative to split BF16, the prefix KV tensors shrink from
1,536 to 969~MiB ($-36.9\%$) and steady device usage falls by 318~MiB
($-7.4\%$). The preparation phase still holds the original cache, so peak
active PyTorch memory rises slightly, from 7,068 to 7,089~MiB. Over nine
trials per configuration, the median post-prefill preparation takes
101.6~ms for the mixed cache and 1.4~ms for split BF16. The mixed
configuration lowers step latency by 9.3\% compared with Flash BF16 but
increases it by 7.5\% relative to split BF16; the matched BF16 kernel is
faster at all five shapes. The 16,384-token measurements concern runtime
and memory only; task quality was measured on shorter contexts.

\section{Related work and scope}\label{sec:related}

Bohbot et al.~\cite{bohbot} (Theorem~5.1) give a high-probability upper
bound on the pooled Lipschitz error for sub-Gaussian tokens, with exponent
$1/(2(1+32H^2))$; their bound for the Gaussian mean has exponent
$1/(2(1+H^2))$. They list sharp finite-temperature Gaussian rates as an open
direction. Their Proposition~5.3 computes the scalar full-hardmax error when
the key coordinate also serves as the value, and Theorem~\ref{thm:aligned}
treats its causal counterpart. Boursier and Boyer~\cite{boursier} analyze
the population limit, local concentration and Jacobians of softmax
attention, with applications to in-context linear regression. For a fixed
query, the partition function is a Gaussian random-energy
sum~\cite{rem}, and the Gaussian tilt is an instance of self-normalized
importance sampling~\cite{importance}. Beyond these single-query analyses,
the main new optimizations here are over the Gaussian query distribution
and, for the pooled second moment, over a key shared by two rows.

Hayase and Karakida~\cite{hayase} derive the scales at which a row
concentrates from the gaps to the largest logit, and Giorlandino and
Goldt~\cite{goldt} compute the participation of a row in their
random-energy analysis of transformers at initialization. We use the
Gaussian one-row freezing scale and participation formula. For pooling,
however, all rows share the same keys, and the probability that two rows
put weight on the same key is what produces the score-rank factor in
Theorem~\ref{thm:growing}. Counting extreme Gaussian points has classical
counterparts in the geometry of Gaussian polytopes~\cite{polytope}, and
concurrence probabilities describe when several maxima share an
index~\cite{concurrence}. Mroueh et al.~\cite{maxout} use the probability
that two fixed directions share a maximizing Gaussian projection as the
locality kernel of random maxout features, and Andoni
et al.~\cite{angularlsh} identify the fixed-angle collision exponent of an
extreme signed Gaussian coordinate in cross-polytope hashing. H\"usler and
Reiss~\cite{husler} give the extreme-value limit for nearly collinear
Gaussian vectors. At equal prefix lengths, our conditional hardmax kernel is
the Brown--Resnick concurrence formula of \cite[Example~6]{concurrence}.
What is new here is the average over anisotropic Gaussian query directions
and unequal causal prefixes; the spectrum-uniform bounds also cover soft
weights and query radii that become rare below the freezing scale.

\paragraph{Scope.} All results concern a single attention layer with
Gaussian tokens in fixed dimension. Growing dimension, learned dependence
between tokens and propagation across layers would require different
estimates.

\paragraph{KV-cache compression.} MiKV and ZipCache already allocate
precision according to token importance, and KIVI quantizes values per
token~\cite{mikv,zipcache,kivi}. QAQ derives the squared-attention
contribution of independent value errors for a single
query~\cite[Sec.~4.1]{qaq}, and the subsequently withdrawn HeadQ preprint
records the same diagonal distortion for fixed attention~\cite[Sec.~3.2]{headq}.
RDKV weights each value token by its cumulative attention, allocates
discrete bit widths from a distortion table and implements packed
decoding~\cite[Eqs.~(2), (6) and Sec.~3.3]{rdkv}. RateQuant allocates bits
by weighted marginal distortion at the granularity of
heads~\cite[Sec.~3.3]{rateq}. ReadKV sums squared attention over the query
heads that share a KV head and uses calibrated marginal gains to decide how
many stored bits to fetch for the current query~\cite[Eq.~(3) and
Sec.~3.3]{readkv}. Attention-weighted value projections optimize a
fixed-weight output objective across the sharing heads~\cite{salfati}, and
Agrawal and Mayer~\cite{servebench} compare task quality and serving cost
across KV-compression systems. Our experiment tests a narrower decision: a
fixed storage precision for each prefix value, chosen from a temporal mean
of attention and the realized reconstruction errors of individual value
vectors. Its score controls isolate individual ranking principles and do not
implement the full codecs of these papers.

\paragraph{Code and data.}
The Gaussian calculation scripts, data for Figure~\ref{fig:spectrum}, and a
Chinese companion with detailed proofs are available at
\href{https://github.com/20000yshust/tcs-gaussian-attention/tree/a0dafcfa5b65a953ea3c97d29cdc1d3f85c15c35}{\nolinkurl{github.com/20000yshust/tcs-gaussian-attention}}
(commit \texttt{a0dafcf}).
The \href{https://github.com/20000yshust/tcs-gaussian-attention/tree/a0dafcfa5b65a953ea3c97d29cdc1d3f85c15c35/experiments/value-cache}{value-cache supplement}
contains the evaluation code, dataset revisions, paired outcomes and timing
samples.

\begingroup
\footnotesize

\small
\noindent\textbf{AI Disclosure Statement.}
OpenAI Codex assisted the author in formulating the mathematical claims,
deriving and writing the proofs, searching the literature, preparing the
manuscripts and figures, and implementing and running the computational
checks and the cache experiments. Exact permutation averages, numerical
minimization and Gaussian quadrature were used to check specific formulas;
the appendices give the analytic arguments for all asymptotic statements.
\endgroup
\clearpage
\appendix
\section{Two estimates for Gaussian samples}

Throughout the appendices we put $\ell=\log n$. For $b\ge0$ define
\[
 f(b)=\max_{0\le s\le1}(1-s^2+2bs)=
 \begin{cases}1+b^2 & 0\le b\le1,\\2b & b\ge1.\end{cases}
\]
This is the free energy of the Gaussian random energy model in our
normalization: for a unit vector $u$ and $t=b\sqrt{2\ell}$, the partition
function $\sum_{j\le n}e^{tu^\top X_j}$ equals $n^{f(b)+o(1)}$ with high
probability. The two lemmas below make this precise in the forms needed
both for fixed queries and for queries drawn from the context.

\begin{lemma}[Empirical Laplace principle]\label{lem:laplace}
Let $G_1,\ldots,G_n\sim\mathcal N(0,I_d)$ be independent and
$g_j=G_j/\sqrt{2\ell}$. Let $\Phi$ be a family of uniformly bounded and
uniformly equicontinuous functions on the closed ball of radius $2$; their
values outside this ball are irrelevant. Then, for each fixed $\eta>0$,
with probability tending to one,
\[
 \left|\log_n\left(\frac1n\sum_j n^{\phi(g_j)}\right)
 -\max_{\norm{z}\le1}\{\phi(z)-\norm{z}^2\}\right|\le\eta
 \qquad\text{simultaneously for all }\phi\in\Phi.
\]
More precisely, the failure probability is $O(n^{-c})$ for some
$c=c(\eta)>0$.
\end{lemma}
\begin{proof}
The density of $g_j$ is $(\ell/\pi)^{d/2} n^{-\norm{z}^2}$. Fix $\delta>0$.
All $g_j$ lie within radius $1+\delta$, except with probability
$n^{1-(1+\delta)^2+o(1)}$.

\emph{Upper bound.} Cover the ball of radius $1+\delta$ by finitely many
cells of diameter at most $\delta$. For each fixed $\eta>0$, Markov's
inequality shows that a cell $D$ contains at most
$n^{1-\inf_{z\in D}\norm{z}^2+\eta}$ points, except with probability
$n^{-\eta+o(1)}$. Cells that leave the unit ball lie within distance
$2\delta$ of it, and projecting them onto the unit ball changes the
objective by a uniform $o_\delta(1)$. Summing over the cells proves the
upper bound up to an error $\eta+o_\delta(1)$ in the exponent.

\emph{Lower bound.} Take a finite $\delta$-net of the ball of radius
$1-2\delta$. The $\delta$-neighborhood $D$ of each center has expected count
at least $n^{1-\sup_{z\in D}\norm{z}^2+o(1)}$, which tends to infinity. By a
binomial Chernoff bound, all these neighborhoods simultaneously contain at
least half of their expected counts, except with superpolynomially small
probability. Every point of the unit ball lies within $3\delta$ of a
center, so uniform equicontinuity gives the lower bound up to an error
$o_\delta(1)$. Letting $\delta$ and $\eta$ tend to zero completes the
proof.
\end{proof}

\begin{lemma}[Lower tail of the denominator]\label{lem:den}
Fix $\bar b<\infty$ and $\eta>0$, let $G_1,G_2,\ldots$ be independent
standard normal variables, and let $m=n-O(1)$. There are constants $c>0$
and $n_0$ such that, for all $n\ge n_0$ and all $b\in[0,\bar b]$,
\[
 \Pp\left\{ \sum_{j=1}^{m} e^{b\sqrt{2\ell}G_j}
                          < n^{f(b)-\eta}\right\}\le e^{-n^c}.
\]
For $b\le1$ the same sum is at most $n^{f(b)+\eta}$ with probability at
least $1-n^{-\eta}$.
\end{lemma}
\begin{proof}
Choose $\delta>0$ small in terms of $\eta$ and $\bar b$, put
$a=\min\{b,1-2\delta\}$, and count the indices $j$ for which
$G_j/\sqrt{2\ell}$ lies within $\delta$ of $a$. The expected count is at
least $n^{1-(a+\delta)^2+o(1)}$, uniformly in $b$. On the event that at least
half of this expectation is realized, these indices contribute at least
$n^{1-(a+\delta)^2+2b(a-\delta)+o(1)}$ to the sum, and this exponent differs
from $f(b)$ by $O((1+\bar b)\delta)$. The claimed tail bound now follows
from the binomial Chernoff bound. For $b\le1$ the expected sum is at most
$n^{1+b^2}$, and Markov's inequality gives the upper bound.
\end{proof}

\section{An individual prediction}

This appendix proves Theorems~\ref{thm:row} and~\ref{thm:critical},
together with the conditional estimate \eqref{eq:conditional}, which is
also used for Theorem~\ref{thm:lipschitz}. We first treat a fixed query
that is independent of the context, then account for the query's own key,
and finally integrate over the Gaussian query distribution.

\paragraph{A fixed query.}
Consider an independent context of $n-O(1)$ tokens and a fixed query $x$.
Write $Ax=t u$ with $t\ge0$ and $\norm{u}=1$ (any unit vector $u$ if
$t=0$), and put $b=t/\sqrt{2\ell}$. Let $\widehat F$ be the prediction
from this context. The error $\widehat F-F(x)$ is the centered numerator
$\sum_j\xi_j$ divided by the denominator $\sum_je^{tu^\top X_j}$, where
the summands
\[
 \xi_j=e^{t u^\top X_j} V(X_j-tu),\qquad \E \xi_j=0,
\]
are independent. Define the conditional error exponents
\[
\begin{split}
r_1(b)&=\begin{cases}
 \tfrac12-b^2&0\le b\le \tfrac12,\\
 (1-b)^2&\tfrac12\le b\le1,\\
 0& b\ge1,
\end{cases}\\[1ex]
r_2(b)&=\begin{cases}
 1-2b^2&0\le b\le1/\sqrt3,\\
 \dfrac{(1-b^2)^2}{4b^2}&1/\sqrt3\le b\le1,\\
 0&b\ge1 .
\end{cases}
\end{split}
\]
For each $p=1,2$, each fixed $\bar b$ and every $\eta>0$, uniformly in
$0\le b\le \bar b$,
\begin{equation}\label{eq:conditional}
 \E\big[\norm{\widehat F-F(x)}^p\mid x\big]
                         \le n^{-r_p(b)+\eta}
 \quad\text{for all sufficiently large } n .
\end{equation}

\emph{Proof of \eqref{eq:conditional}.} Whatever the weights, the error is
at most $\norm{V}_{\mathrm{op}}(\max_j\norm{X_j}+t)$. By the Cauchy--Schwarz
inequality and Gaussian bounds on the moments of the maximum, the
exceptional event of Lemma~\ref{lem:den} therefore contributes negligibly
to either error moment. On its complement the denominator contributes the
factor $n^{-p f(b)+p\delta}$, where the slack $\delta>0$ can be chosen
arbitrarily small. Write $S=u^\top X$. For any fixed bounded $b$ and $c$,
the Gaussian tilt identity yields
\begin{align}
 \E[\norm{\xi_j}^2;\ S\le c\sqrt{2\ell}]
 &\le n^{4b^2-(2b-c)_+^2+o(1)},\label{eq:tilt2}\\
 \E[\norm{\xi_j};\ S>c\sqrt{2\ell}]
 &\le n^{b^2-(c-b)_+^2+o(1)}.\label{eq:tilt1}
\end{align}
Indeed, tilting by $e^{2tS}$ and by $e^{tS}$ shifts the mean of $S$ by $2t$
and $t$, respectively. Integrating the remaining factors, of degree at most
two, against the shifted Gaussian tails costs at most a polynomial factor
in $\ell$, and the coordinates orthogonal to $u$ have fixed Gaussian
moments. For the untruncated second moment, the tail penalty in
\eqref{eq:tilt2} is simply omitted.

\emph{First moment ($p=1$).} For $b\le1/2$, the numerator has mean zero, so
its expected norm is at most $\sqrt{n\,\E\norm{\xi_j}^2}$; after division by
the denominator this gives $n^{-1/2+b^2+o(1)}$. For $1/2\le b\le1$,
truncate at $c=1$ and center the truncated summands. The expected norm of
the centered part is still at most the square root of the summed
variances, while the residual and the centering correction are together at
most twice the sum of the first moments in \eqref{eq:tilt1}. Both bounds
are $n^{2b+o(1)}$ before division, which gives $n^{-(1-b)^2+o(1)}$.

\emph{Second moment ($p=2$).} For $b\le1/\sqrt3$, the full variance gives
$n^{1+4b^2-2(1+b^2)+o(1)}$. For $1/\sqrt3\le b\le1$, use the cutoff
$c=(1+b^2)/(2b)$, the projection at which a single key carries a weight
comparable to the typical denominator. The probability that some score
exceeds $c\sqrt{2\ell}$ is at most $n^{1-c^2+o(1)}$. On the event
$\max_j\norm{X_j}\le M\sqrt{2\ell}$, for a large constant $M$, the squared
error is $O(\ell)$, and Gaussian tail integration makes the complement of
this event negligible, uniformly in $b$. On the event that no score exceeds
the cutoff, apply \eqref{eq:tilt2} to the variance of the truncated
numerator and \eqref{eq:tilt1} to its bias. After normalization these two
terms have exponents $1-c^2$ and $-2(c-b)^2=2(1-c^2)$, respectively.
Finally, for $b\ge1$ the bound by the maximum alone suffices, because
$r_p(b)=0$ there.

\paragraph{Including the query among the keys.}
Now let the query also serve as a key. Write $x=\sqrt{2\ell}\,z$ and let
$\lambda$ be the weight of this self key, so that the error becomes
$(1-\lambda)(\widehat F-F(x))+\lambda V(x-Ax)$. By Lemma~\ref{lem:den}, up
to an arbitrarily small slack in the exponent,
$\lambda\le\min(1,n^{2z^\top A z-f(\norm{Az})})$ on bounded sets of $z$. If
$s=\norm{z}\le1$ and $b=\norm{Az}$, the variational formula for $f$, taken
at this $s$, gives $f(b)\ge 1-s^2+2bs\ge1-s^2+2z^\top A z$. Consequently
\[
 \E[\lambda^p(1+\norm{x})^p]\le n^{-1+o(1)},\qquad p=1,2.
\]
To see this, on $\{s\le1\}$ combine the query density, whose exponent is
$s^2$, with the weight bound, whose exponent is $p(1-s^2)$; their sum is at
least one. On $\{s>1\}$ use $\lambda\le1$ and the Gaussian tail of the
query. Failures of the denominator bound on bounded sets are negligible.

\paragraph{Integration over queries.}
Integrating \eqref{eq:conditional} over the query gives the exponents
$\inf_{z\in\R^d}\{\norm{z}^2+r_p(\norm{Az})\}$, and by the preceding
paragraph the same bounds hold for self-attention. For the upper bound,
integrate first over a fixed large ball, using the scaled Gaussian density;
by the bound on the maximal error, the contribution from outside the ball
decays faster. For $H>0$ the infimum equals
$\inf_{b\ge0}\{b^2/H^2+r_p(b)\}$, since equality in $\norm{Az}\le H\norm{z}$
is attained along a top right singular direction. For $H=0$ it equals
$r_p(0)$.

For $p=1$, the branch $b\le1/2$ is minimized at $b=0$ if $H\le1$ and at
$b=1/2$ otherwise, and on $[1/2,1]$ the unconstrained minimizer is
$b=H^2/(1+H^2)$. Together these give $\beta(H)$. For $p=2$ write $v=b^2$.
For $v\in[0,1/3]$ the objective is $1+(H^{-2}-2)v$, and for $v\in[1/3,1]$ it
is
\[
 \frac1{4v}-\frac12+\left(\frac14+H^{-2}\right)v .
\]
The unconstrained minimizer $v_*=H/\sqrt{H^2+4}$ of the latter lies in this
branch exactly when $H\ge1/\sqrt2$, and the resulting infimum is
$\gamma_1(H)$. In both computations the region $b\ge1$ cannot improve the
infimum, because there the objective is smallest at $b=1$.

\paragraph{Matching lower bound for the squared error.}
For any fixed query $x$, the multivariate central limit theorem for the
centered numerator, combined with the law of large numbers for the
denominator, shows that the $\sqrt n$-scaled error has a nonzero limiting
covariance whenever $V\ne0$. For a fixed query the self contribution
vanishes on this scale in probability. Lower semicontinuity of the second
moment under convergence in law, together with Fatou's lemma, then gives
the global lower bound $\Omega(n^{-1})$.

For $H>1/\sqrt2$, take $z_*$ in a top right singular direction with
$\norm{Az_*}^2=v_*$; then $\norm{z_*}^2=v_*/H^2=1/(H\sqrt{H^2+4})<1$.
Write $c(b)=(1+b^2)/(2b)$ for the cutoff used in the upper bound, and
restrict $z$ to a small fixed neighborhood of $z_*$. In this neighborhood $1/\sqrt3<b<1$,
$c(b)>1$, and the exponent $2z^\top A z$ of the self score is strictly
below $f(b)$. Fix a small $\delta>0$. With conditional probability at
least $n^{1-(c(b)+2\delta)^2+o(1)}$, exactly one of the independent keys has
$S/\sqrt{2\ell}\in[c(b)+\delta,c(b)+2\delta]$, all other projections are
bounded in absolute value by $(c(b)+\delta)\sqrt{2\ell}$, and the partition
sum of the other keys is at most $n^{f(b)+b\delta}$. For a specified
outlier, the conditions on the other keys hold with probability
$1-o(1)$ by Lemma~\ref{lem:den} and a Gaussian union bound, and the events
for different outlier indices are disjoint. Since $2bc(b)=f(b)$, the
exponent of the outlier exceeds that of the other keys' sum by at least
$2b\delta-b\delta=b\delta>0$. The self exponent stays strictly below $f(b)$
on a sufficiently small closed neighborhood, so the weight of the outlier
tends to one uniformly there.

Now take any nonzero row $v^\top$ of $V$. Given all projections, the
orthogonal parts of the independent keys remain Gaussian. If
$|v^\top u|\le\norm{v}/2$, the conditional variance of this output
coordinate is bounded below by a positive constant on the outlier event.
If $|v^\top u|>\norm{v}/2$, the conditional mean differs from the
population coordinate by order $\sqrt{\ell}$. Indeed, the outlier carries
weight $1-o(1)$, and its projection exceeds the target $b\sqrt{2\ell}$ by at
least $(c(b)-b)\sqrt{2\ell}$, where $c(b)-b=(1-b^2)/(2b)$ is bounded away
from zero on this neighborhood; the remaining projected values and the self
term contribute only $o(\sqrt{\ell})$ to the difference. In either case the conditional squared error is bounded below by a positive
constant on the outlier event. Integrating over the query neighborhood,
then shrinking it and letting $\delta\downarrow0$, gives the lower exponent
$\norm{z_*}^2+c(\norm{Az_*})^2-1$. Since $c(b)^2-1=(1-v)^2/(4v)$ with
$v=b^2$, this is the value of the objective
$1/(4v)-1/2+(1/4+H^{-2})v$ at $v_*$, namely $\gamma_1(H)$. This proves
Theorem~\ref{thm:row}.

\paragraph{The leading constant and the critical logarithm.}
We now prove Theorem~\ref{thm:critical}. First, for any $b_0<1$ and
uniformly in $b=\norm{Ax}/\sqrt{2\ell}\le b_0$, the denominator $D$ of an
independent context satisfies
\[
 \Pp\{D<\tfrac12 m_n e^{t^2/2}\}\le e^{-n^c},
 \qquad m_n=n-O(1),
\]
for some $c>0$. To verify this, truncate the weights at
$S\le (b+\delta)\sqrt{2\ell}$ for a fixed, sufficiently small $\delta>0$.
By exponential tilting the retained mean is $(1-o(1))e^{t^2/2}$, and each
retained weight is at most $n^{2b(b+\delta)}$. After the weights are scaled
into $[0,1]$, a multiplicative lower-tail Chernoff bound has an exponent of
at least a constant times $n^{1-b^2-2b\delta}$, which is a positive power
of $n$ uniformly in $b\le b_0$. On this good denominator event we use the
full second moment of the centered numerator. The conditional risk for a
fixed query is therefore at most
\begin{equation}\label{eq:relative-variance}
 \frac{C}{n} e^{t^2}\big(\norm{V}_{\mathrm F}^2+\norm{Vtu}^2\big)
 +e^{-n^{c' }}
\end{equation}
uniformly on this range; on the bad event we use the Gaussian bound on the
maximum, as before.

For $H<1/\sqrt2$, choose $b_0$ with $H<b_0<1$. For $H>0$ the query tail
$\{b>b_0\}$ contributes $n^{-b_0^2/H^2+o(1)}=o(1/n)$; at $H=0$ the attention
is uniform. For a fixed $x$, the normalized denominator tends to one in
probability, and the squared norm of the centered numerator divided by $n$
is uniformly integrable, because its fourth moment is bounded on this scale.
Together with the lower tail of the denominator, this gives the
conditional limit $n R_n(x)\to e^{t^2}(\norm{V}_{\mathrm F}^2+\norm{Vtu}^2)$.
(The limit can also be read off directly by applying the Gaussian tilt with
parameter $2tu$ to the variance of a centered summand.) The right-hand side
is integrable in $x$ since $H^2<1/2$, so \eqref{eq:relative-variance} and
dominated convergence give the asserted constant.

At $H=1/\sqrt2$, the Gaussian density multiplied by $e^{\norm{Ax}^2}$ is
proportional to $e^{-x^\top(I/2-A^\top A)x}$: it is constant along
$\mathcal E$ and decays as a nondegenerate Gaussian on $\mathcal E^\perp$.
Choose $b_0$ with $H<b_0<1$. On $\{\norm{Ax}\le b_0\sqrt{2\ell}\}$ the
radius in $\mathcal E$ is $O(\sqrt{\ell})$, so the integral of the factor
in parentheses in \eqref{eq:relative-variance} is $O(\ell^{\mu/2+\iota})$.
For a matching lower bound it suffices to consider $b\le1/4$. Write $U=\sum_{j=1}^{m_n}\xi_j$.
Its second moment is $m_n e^{2t^2}(\norm{V}_{\mathrm F}^2+\norm{Vtu}^2)$,
and independence together with a tilt at $4tu$ gives
\[
 \frac{\E\norm{U}^4}{(\E\norm{U}^2)^2}
 \le C(1+n^{-1}e^{4t^2})=O(1).
\]
Here the fourth moment of the sum is bounded by a constant times the sum
of the fourth moments plus the square of the sum of the second moments, and
the tilted fourth moment of a single summand is at most
$C e^{8t^2}(\norm{V}_{\mathrm F}^2+\norm{Vtu}^2)^2$. By Chebyshev's
inequality the event $D>2m_n e^{t^2/2}$ has probability
$O(n^{-1}e^{t^2})=o(1)$, and the Cauchy--Schwarz inequality shows that it
carries only an $o(1)$ fraction of the second moment of the numerator.
Hence the conditional risk is bounded below by a constant times the first
term of \eqref{eq:relative-variance}. Integrate over the region
$\norm{x_{\mathcal E}}\le c_0\sqrt{\ell}$,
$\norm{x_{\mathcal E^\perp}}\le1$, for a small fixed $c_0>0$; this region
satisfies $b\le1/4$ for large $n$. The constant term in parentheses supplies
the factor $\ell^{\mu/2}$, and when $VA|_{\mathcal E}\ne0$ the squared
linear term supplies $\ell^{\mu/2+1}$, by the symmetry of the integration
over $\mathcal E$.

For $H\le1/\sqrt2$, adding the self key changes the output by $o(n^{-1/2})$
in quadratic mean. In fact, whenever $s=\norm{z}\le1/H$ one has
$b\le Hs\le1$, and since $f(b)=1+b^2$ and $2z^\top Az\le2sb$, the bound on
the self score improves to
$f(b)-2z^\top Az\ge1-s^2+(s-b)^2\ge1-s^2+(1-H)^2 s^2$. Hence
$s^2+2(f(b)-2z^\top Az)_+\ge 2-s^2+2(1-H)^2s^2\ge\min\{2,1+2(1-H)^2\}$ on
$\{s\le1\}$, and by continuity this exponent stays uniformly above one on
$\{1\le s\le1+\delta\}$ for some fixed small $\delta>0$. Above $1+\delta$
the query density alone contributes an exponent greater than one.
Lemma~\ref{lem:den} and the Gaussian bound on the maximum therefore show
that the expected squared change is $o(1/n)$,
and the triangle inequality in $L^2$ transfers both results to
self-attention. For the causal average, the sum of $(\log i)^a/i$ up to $n$
is of order $(\log n)^{a+1}$ for $a\ge0$, which gives the claimed
logarithms.

\section{Pooled statistics and typical errors}
\label{sec:typical}

\paragraph{The upper bound in Theorem~\ref{thm:lipschitz}.}
The difference $L_n(h)$ is at most the mean of $\norm{F_n(X_i)-F(X_i)}$
plus the absolute value of the centered sample average of $h(F(X_i))$. By
exchangeability and \eqref{eq:conditional}, the first term has expectation
at most $n^{-\beta(H)+o(1)}$. The expectation of the second is at most
$n^{-1/2}\sqrt{\operatorname{Var}(h(F(X)))}$, which is bounded uniformly in
$1$-Lipschitz $h$ because one can subtract $h(0)$. This proves the stated
upper bound.

\paragraph{Typical row errors in the marked head.}
For the matching examples and for the typical errors we use the marked head
of Theorem~\ref{thm:marked}. Write $w_{ij}$ for the attention matrix,
$\rho_i=\sum_j w_{ij}^2$ for the participation of row $i$, and
$p_j=n^{-1}\sum_i w_{ij}$ for the mass of column $j$. Conditional on the
$G_i$, each $Y_i=\sum_j w_{ij}Z_j$ is centered Gaussian with variance
$\rho_i$, and $\overline Y$ has variance $\sum_j p_j^2$. Set
$x_i=G_i/\sqrt{2\ell}$ and $b_i=\norm{Ax_i}$; from now on the letters
$x,y,z$ denote such scaled vectors. Applying
Lemma~\ref{lem:laplace} to the partition functions at inverse temperatures
$1$ and $2$ gives
\[
 \rho_i=n^{f(2b_i)-2f(b_i)+o_{\mathbb P}(1)}
\]
uniformly in $i$, because the lemma controls all query vectors in a fixed
ball of scaled coordinates simultaneously. One more application of the
lemma, now over the queries, yields, for $a=1$ and $a=1/2$,
\begin{equation}\label{eq:empirical}
 \frac1n\sum_i \rho_i^a
 =
 n^{-\min_{0\le b\le H}\{b^2/H^2+a(2 f(b)-f(2b))\}
       +o_{\mathbb P}(1)}
\end{equation}
(at $H=0$, use the single value $b=0$). The function $2 f(b)-f(2b)$ equals
$1-2b^2$ for $b\le1/2$, $2(1-b)^2$ for $1/2\le b\le1$, and zero for $b\ge1$.
For $a=1$ the minimum is $\min\{1,2/(1+2H^2)\}$, and for $a=1/2$ it is
$\beta(H)$. In their respective nonconstant regimes the two minimizers are
$b=2H^2/(1+2H^2)$ and $b=H^2/(1+H^2)$, both at most $H$.

\paragraph{From conditional variances to the observed errors.}
The random values preserve these exponents. Conditional on the scores,
$Q_n$ is distributed as $\sum_j\lambda_j N_j^2$, where the $N_j$ are
independent standard normal variables, $\lambda_j\ge0$ and
$\sum_j\lambda_j=n^{-1}\sum_i\rho_i$. Markov's inequality controls large
values of the ratio of $Q_n$ to this trace. For small values, the Laplace
transform of independent squared normals, together with
$\prod_j(1+2\lambda_j/(\epsilon\sum_j\lambda_j))\ge1+2/\epsilon$, shows that
the ratio is at most $\epsilon$ with probability at most
$e(1+2/\epsilon)^{-1/2}$. Similarly, let $M_n=n^{-1}\sum_i |Y_i|$, which
is exactly $L_n(|\cdot|)$ in this model, and divide it by
$n^{-1}\sum_i\sqrt{\rho_i}$. The ratio is a weighted mean of absolute
standard normal variables, which may be dependent. If the ratio is at most
$\epsilon$, then the coordinates whose absolute value is at most
$\sqrt{\epsilon}$ carry total weight at least $1-\sqrt{\epsilon}$. The
expected total weight of those coordinates is $O(\sqrt{\epsilon})$,
uniformly in the weights and the covariance, so Markov's inequality bounds
the probability of a small ratio by $O(\sqrt\epsilon)$; large values are
again controlled by Markov's inequality. This proves the claimed typical
exponents. The lower bound on $\E M_n$ follows directly from the typical
lower bound, and its upper bound was proved above. Since a single score
coordinate ($d=1$) already attains these exponents, the extremal head needs
only two coordinates.

\paragraph{The typical pooled mean.}
To evaluate the typical pooled mean, apply Lemma~\ref{lem:laplace} to the
queries contributing to each column. For $\norm{z}\le1$, the exponent of
$p_j$ at $x_j=z$ is
\[
 \kappa(z)=\min_{\norm{x}\le1}
 \{\norm{x}^2+ f(\norm{Ax})-2 z^\top A x\}.
\]
On the branch $\norm{Ax}\le1$ the quadratic minimizer is
$x_*=(I+A^\top A)^{-1}A^\top z$, which has norm at most $1/2$ and satisfies
$\norm{Ax_*}\le H^2/(1+H^2)<1$. On the other branch the objective is
nondecreasing along every ray, since $z^\top Ax\le\norm{Ax}$. Thus
\[
 \kappa(z)=1-z^\top A(I+A^\top A)^{-1}A^\top z .
\]
The variational expression is continuous on bounded sets, so this formula
also controls sample points within $o_{\mathbb P}(1)$ of the unit ball.
The largest eigenvalue of the matrix $A(I+A^\top A)^{-1}A^\top$ is
$H^2/(1+H^2)$, and a final application of Lemma~\ref{lem:laplace} gives
$\sum_j p_j^2=n^{-\min\{1,2/(1+H^2)\}+o_{\mathbb P}(1)}$. Conditional on
the score coordinates, $P_n$ is this variance times a chi-square variable
with one degree of freedom, whose upper and lower tails on polynomial
scales preserve the exponent.

\section{Expected squared error of the pooled mean}
\label{sec:expected-pool}

For independent value coordinates, exchangeability gives
\[
 \E P_n=\E\sum_j p_j^2=\frac1n\sum_{i,k}\E[w_{i1}w_{k1}] .
\]
We first dispose of the terms in which two of the three indices coincide.
The terms with $i=k\ne1$ contribute at most $n^{-1-\gamma_1(H)+o(1)}$ in
total, since $\E\sum_jw_{2j}^2=\E Q_n$ and the columns other than the
query's own are exchangeable. In the terms where exactly one of $i,k$
equals $1$, the Cauchy--Schwarz inequality and the bound on the self
weight give $n^{-1-\gamma_1(H)/2+o(1)}$. The term with $i=k=1$ is
$n^{-2+o(1)}$. It remains to estimate $n\E[w_{21}w_{31}]$. For $H=0$ the
weights are uniform and $\E P_n=1/n$.

\paragraph{Upper bound.}
Condition on the three scaled vectors $(z,x,y)=(x_1,x_2,x_3)$. Dropping
these keys from the denominator and applying Lemma~\ref{lem:den}, we obtain
on bounded sets the estimate
\[
 w_{21}\le n^{-(f(\norm{Ax})-2 z^\top Ax)_++\eta}
\]
outside an event of negligible conditional probability, and the analogous
bound for $w_{31}$. On the complement of a large bounded set the Gaussian
tail alone suffices. Therefore the exponent of $n\E[w_{21}w_{31}]$ is at
least
\[
 \inf_{z,x,y}\left\{\norm{z}^2+\norm{x}^2+\norm{y}^2-1+
 (f(\norm{Ax})-2z^\top Ax)_++
 (f(\norm{Ay})-2z^\top Ay)_+\right\}.
\]
Putting $s=\norm{z}$ reduces this to
\[
 \inf_{s\ge0}\{s^2-1+2g(s)\},\qquad
 g(s)=\min_{b\ge0}\{b^2/H^2+(f(b)-2sb)_+\}.
\]
The reduction is exact: take $z$ along a top left singular vector and both
queries along the corresponding right singular vector. Let
$\alpha=H^2/(1+H^2)$. For $s\le1$ the optimum is $b=\alpha s$. For $s>1$
the penalty vanishes at and above $b_0=s-\sqrt{s^2-1}$, and the minimizing
$b$ is $\min(\alpha s,b_0)$; this follows by minimizing
$1+(1+H^{-2})b^2-2sb$ below $b_0$ and then $b^2/H^2$ above it. For $s\le1$
the region $b\ge1$ cannot improve the optimum, because the objective is
nondecreasing there. The two regimes switch at
$s_0=1/\sqrt{\alpha(2-\alpha)}$. For $s\le s_0$ the outer objective is
$1+(1-2\alpha)s^2$. For $s\ge s_0$ set $v=b_0^2$; the outer objective then
becomes
\[
 \frac1{4v}-\frac12+\left(\frac14+\frac{2}{H^2}\right)v,
 \qquad 0<v\le\frac{H^2}{H^2+2}.
\]
If $H\le1$ the global minimum is $1$. If $H>1$, the unconstrained minimizer
$v=H/\sqrt{H^2+8}$ lies in the second interval, and the minimum is
$\gamma_2(H)$.

\paragraph{Lower bound.}
When $H\le1$, the bound $\sum_j p_j^2\ge1/n$ suffices. When $H>1$, take
aligned $z,x,y$ at the minimizing triple, with
$\norm{Ax}=\norm{Ay}=\sqrt v<1$, $\norm{x}=\norm{y}=\sqrt v/H<1$ and
$2z^\top Ax=2z^\top Ay=f(\sqrt v)$, and restrict to a small fixed
neighborhood of this triple. The shared key then has norm
$\norm{z}=(1+v)/(2\sqrt v)>1$: the expectation is driven by a rare key
outside the unit ball that captures a weight $n^{-o(1)}$ in both rows. In
each of the two rows, the scores of the keys $x$ and $y$ have exponents
strictly below the free energy of the denominator, because their norms are
less than one. By
Lemma~\ref{lem:den}, the remaining partition sums are bounded above with
probability $1-o(1)$, simultaneously for both queries. If the
neighborhood has radius $\delta$ and the slack is chosen accordingly, the
distinguished key $z$ therefore has weight at least $n^{-o_\delta(1)}$ in
both rows. Integrating its Gaussian probability and letting $\delta$ tend
to zero gives the matching exponent for $n\E[w_{21}w_{31}]$, which
completes the proof of Theorem~\ref{thm:marked}.

\section{Causal self-attention}

The last token sees all $n$ keys, and the expected row errors at position
$i$ are those for a context of size $i$. Averaging the expectation bounds
over positions uses $n^{-1}\sum_{i=1}^n i^{-\gamma+o(1)}=n^{-\gamma+o(1)}$
for $0<\gamma\le1$, provided matching lower bounds hold: the last half of
the indices supplies the lower bound, and the harmonic sum handles
$\gamma=1$. The same upper-bound argument applies to pooled Lipschitz
observables, with $\gamma=\beta(H)$. For example, in the marked head at
$H=0$ the uniform prefix weights give $\E Q_n=h_n/n$ and
$\E P_n=(2n-h_n)/n^2$, where $h_n=\sum_{i=1}^n 1/i$; for the second
identity, note that the covariance of the prefix averages at positions $i$
and $k$ is $1/\max(i,k)$. Under full attention both risks equal $1/n$ at
$H=0$.

\paragraph{Typical errors.}
For the remaining claims we group the queries into dyadic blocks
$I_t=\{t+1,\ldots,2t\}$, $t=1,2,4,\ldots$, and treat the first row as a
separate summand. For upper bounds, an incomplete final block may be
completed with fresh tokens, since this comparison involves only
nonnegative weights, column masses or powers of $\rho_i$; the conditional
Gaussian step is applied to the actual rows afterwards. Within a block,
each denominator lies between the sums over the first $t$ and over the
first $2t$ keys, and the same holds for the numerator of $\rho_i$ at twice
the temperature. With scaled tokens $G_j/\sqrt{2\log t}$,
Lemma~\ref{lem:laplace} gives the same logarithmic estimate for both of
these key lists, uniformly over the queries of the block. Applying it to
the $t$ query tokens then yields \eqref{eq:empirical} for the mean over
$I_t$, with $t$ in place of $n$; replacing the number of keys $t$ by $2t$
changes a logarithmic exponent only by $o(1)$. For a fixed tolerance these
estimates hold simultaneously for all dyadic $t\ge n^\xi$, for each fixed
$\xi>0$, by the polynomial failure bound in Lemma~\ref{lem:laplace}: the
union of the failure events has probability $O((\log n)n^{-\xi c})$. The
earlier blocks contribute $O(n^\xi/n)$ to the global mean of $\rho_i^a$,
since $\rho_i^a\le1$. Summing over the blocks gives the asserted upper
exponents. In detail, given a target exponent $g\le1$ and a slack
$\eta>0$, choose $0<\xi<\eta$. The early contribution is then
$O(n^{-g+\eta})$, and the later blocks contribute at most
$n^{-1}\sum_t t^{1-g+\eta}=O(n^{-g+\eta})$. A complete dyadic block with
$n/4<t\le n/2$ supplies the lower bound (at an endpoint one may take
$t=n/2$). The conditional Gaussian arguments for $Q_n$ and $M_n$ are
unchanged.

For the typical pooled mean, put $p^{(t)}_j=t^{-1}\sum_{i\in I_t}w_{ij}$,
extended by zeros to all positions. To bound $\norm{p^{(t)}}^2$ from above,
use the first $t$ keys to bound each denominator from below and sum the
resulting bound over all $2t$ columns. For a lower bound, retain only the
first $t$ columns and use the first $2t$ keys in each denominator. By the
column calculation of Section~\ref{sec:typical}, both estimates give the
exponent $\min\{1,2/(1+H^2)\}$ for the squared norm of the block. For the
global upper bound, apply the triangle inequality to
$n^{-1}\sum_t t p^{(t)}$, adding the first row separately; for the lower
bound, use nonnegativity and any full block with $t$ comparable to $n$. The
same simultaneous event and the same bound on early blocks apply as above.

\paragraph{The expected pooled mean.}
For a full block, the expected squared norm is
\[
 \E\norm{p^{(t)}}^2
 =\frac1{t^2}\sum_j\sum_{i,k\in I_t}\E[w_{ij}w_{kj}] .
\]
There are $O(t^3)$ terms with three distinct indices. Conditional on their
tokens, the first $t$ keys, minus at most three distinguished ones, supply
the lower estimate for the denominator, and a term in which $j$ lies
outside a prefix vanishes. Thus the upper-bound calculation of
Section~\ref{sec:expected-pool}, with $t$ as the scale, applies uniformly
to these terms. For the $O(t^2)$ diagonal terms $i=k\ne j$, an off-diagonal
weight in a row of length $i$ has second moment at most
$t^{-1-\gamma_1(H)+o(1)}$ by exchangeability of the past keys. When $j$
equals exactly one of $i,k$, the Cauchy--Schwarz inequality and the
self-weight estimate give $t^{-1-\gamma_1(H)/2+o(1)}$ for each of the
$O(t^2)$ terms. When all three indices coincide, each of the $O(t)$ terms
is at most $t^{-1+o(1)}$. Together these contributions prove
$\E\norm{p^{(t)}}^2\le t^{-\gamma_2(H)+o(1)}$. By Minkowski's inequality in
$L^2(\ell^2)$, the global $L^2$ norm is at most
$n^{-1}+n^{-1}\sum_t t(\E\|p^{(t)}\|^2)^{1/2}$, and the same geometric sum,
now with exponent $\gamma_2(H)/2$, gives the global upper bound. For the
lower bound with $H>1$, take a full block with $t$ comparable to $n$,
distinct queries $i,k\in I_t$, and a key $j\le t$. All $\Theta(t^3)$ such
terms have both weights present, and the lower-bound construction of
Section~\ref{sec:expected-pool} applies uniformly to them, since the first
$2t$ keys supply the upper estimate for the denominator and contain the
same three distinguished keys. Positivity then gives the global lower
bound. For $H\le1$ it suffices that the pooled weights sum to one. This
proves Theorem~\ref{thm:causal}.

\section{Proofs for growing horizons}\label{sec:growing-proof}

\subsection{A single softmax row}

For a weight vector $w$ write $\rho(w)=\sum_j w_j^2$ for its participation.
We first consider $m$ independent standard normal projections $S_j$ and the
weights $w_j(t)\propto e^{tS_j}$ with $t\ge0$, that is, a single softmax row
whose query is independent of the keys. The first paragraph below explains
how rows of self-attention reduce to this situation.

\paragraph{Removing distinguished keys.}
When we study one or two rows, we remove their own query tokens from both
key lists before conditioning on the queries. For a list of original length
$m$, the expected total weight of the removed keys is $O(m^{-\kappa})$ for
some $\kappa>0$, uniformly in the strength. To see this, note that, except
on an event of probability $m^{-1/4+o(1)}$, the at most two distinguished
vectors have norm at most $a_m/2$. Conditional on these vectors, every other projection along
the given unit score direction is standard normal. The number of
projections exceeding $a_m/2$ has mean $m^{3/4+o(1)}$, and a binomial
lower-tail bound shows that it exceeds $m^{2/3}$, except with
superpolynomially small probability. On this event each removed key, whose
projection is at most its norm, is beaten by at least $m^{2/3}$ other keys
and therefore has weight at most $m^{-2/3}$; at zero strength the weights
are uniform. Removing the distinguished keys and renormalizing changes the
weight vector by twice the removed mass in $\ell^1$ (with zeros inserted at
the removed indices). This controls row overlaps, $\rho$, $\sqrt{\rho}$ and
the mass of any subset, all of which are Lipschitz in $\ell^1$ with a fixed
constant. Since the estimate is unconditional, it also bounds the expected
change multiplied by any indicator of the queries.

\paragraph{Diffuse and concentrated rows.}
Rows below the freezing scale are diffuse: for every fixed $\delta>0$ there
is $\kappa_\delta>0$ such that, whenever $0\le t/a_m\le1-\delta$,
\begin{equation}\label{eq:diffuse}
      \E \sqrt{\rho(w(t))}\le m^{-\kappa_\delta}.
\end{equation}
Indeed, the largest projection is at most $(1+\eta)a_m$ except with
polynomially small probability, and by Lemma~\ref{lem:den} the denominator
is at least $m^{1+b^2-\eta}$, where $b=t/a_m$. Thus the largest weight is
at most $m^{-(1-b)^2+(2b+1)\eta}$. It remains to choose $\eta$ small and to
use $\rho\le\max_j w_j\le1$.

Rows above the freezing scale are concentrated. Put $z_n=\Phi^{-1}(1-1/n)$,
where $\Phi$ is the standard normal distribution function and $\phi$ its
density, and write $\zeta_K=z_n-K/z_n$. For fixed $\delta,\epsilon>0$ there
are constants $K,D<\infty$ with the following property: for all large $n$,
all $n/3\le m\le n$ and all $t\ge(1+\delta)a_n$, with probability at least
$1-\epsilon$ the keys with projections above $\zeta_K$ carry total weight at
least $1/2$, and there are at most $D$ such keys. In particular,
$\rho\ge1/(4D)$ on this event.

To prove this, recall from Mills' ratio that $z_n/a_n\to1$ and
$n\phi(z_n)/z_n\to1$. Since $n\overline\Phi(z_n-L/z_n)\to e^L$, we can
choose $L>0$ such that $\max_{j\le m}S_j\ge\zeta_L$ except with
probability $\epsilon/3$. Let $K>L$, so that $\zeta_K<\zeta_L$, and put
$t_0=(1+\delta)a_n$. On the event just described, the fraction of weight
on projections below $\zeta_K$ is at most
$\sum_{S_j<\zeta_K}e^{t_0(S_j-\zeta_L)}$, simultaneously for all $t\ge t_0$.
Integrating the shifted Gaussian density over $(-\infty,\zeta_K]$ shows
that the unconditional expectation of this sum is at most
\[
 \frac{n\phi(\zeta_K)}{t_0-\zeta_K}\, e^{-t_0(\zeta_L-\zeta_K)}
 \le C_{\delta,L}\,e^{-\delta K/2}
\]
for all sufficiently large $n$. We now choose $K$ large and apply Markov's
inequality to this sum and to the number of projections above $\zeta_K$,
whose expectation is at most $2e^K$ for large $n$. This proves both claims.
Increasing $K$ shows in the same way that the high projections can be made
to carry any fixed fraction $1-\epsilon$ of the weight.

\paragraph{The exact limiting row risk.}
For independent keys, Gaussian integration by parts gives
\begin{equation}\label{eq:stein-growing}
 \E\sum_j w_j(t)S_j
   =\sum_j\E\,\partial_{S_j}w_j(t)=t(1-\E\rho(w(t))).
\end{equation}
If $t/a_m\to b\in(0,\infty)$, the left side divided by $a_m$ tends to
$\min(b,1)$. To see this, note that $\sum_jw_jS_j/a_m$ is the derivative at
$y=t/a_m$ of the convex function
$y\mapsto a_m^{-2}\log(m^{-1}\sum_j e^{a_m y S_j})$, and that by
Lemma~\ref{lem:laplace} these functions converge in probability, uniformly
on compact sets, to $(f(|y|)-1)/2$. Bounding the derivative by the two
adjacent secants and then shrinking the secant length gives convergence of
the derivative, including at $b=1$, where the limit is still
differentiable. Gaussian bounds on the maximum give uniform integrability,
and hence $\E\rho(w(t))\to(1-1/b)_+$. For $t/a_m\to0$ we use
\eqref{eq:diffuse}, and for $t/a_m\to\infty$ we use $\E\max_j|S_j|=O(a_m)$
in \eqref{eq:stein-growing}. In the marked head, the query $G_i$ has
strength $t=H_n\|BG_i\|$ along the direction $BG_i/\|BG_i\|$, so
conditioning on $G_i$, dominated convergence and the removal estimate give
the limit $q_c(B)$. For causal prefixes with $i/n\to s>0$ we have
$a_i/a_n\to1$, so the same limit holds. It is uniform over $i\ge\epsilon n$
by a subsequence argument, and the first $\epsilon n$ rows contribute at
most $\epsilon$ to the average. This proves the limit of $\E Q_n$ in
part~3 of Theorem~\ref{thm:growing}.

\subsection{The upper bound from the angle between queries}\label{sec:angle}

Conditional on the score tokens, let $p_j=n^{-1}\sum_i w_{ij}$. Then
$\E_Z P_n=\|p\|^2$ and $\|p\|^2=n^{-2}\sum_{i,k}\langle w_i,w_k\rangle$, so
it suffices to bound the expected overlap of two distinct rows. For two
distinct queries, remove the two distinguished indices, let $u,v$ be the
resulting query directions, and put $s=|u^\top v|$. Order the two rows so
that their reduced lists have lengths $m_i\le m_k$. For a common key $j$,
the two weights are functions $g$ and $h$ of two Gaussian projection arrays
that are independent across keys and have coordinatewise correlation
$u^\top v$; one of the two functions ignores the coordinates outside its
prefix. The Gaussian hypercontractive inequality then gives
\begin{equation}\label{eq:angular-hyper}
 \E[g(X)h(Y)]\le\|g\|_{1+s}\|h\|_{1+s}
     \le(m_i m_k)^{-1/(1+s)} .
\end{equation}
For the first inequality, reflect the second array if the correlation is
negative, and combine H\"older's inequality with the Gaussian
noise-operator bound $\|T_s h\|_{(1+s)/s}\le\|h\|_{1+s}$~\cite{hyper}; at
the endpoints $s=0$ and $s=1$, use independence and the Cauchy--Schwarz
inequality instead. The second inequality uses $0\le g,h\le1$ and,
conditional on the queries, exchangeability within each reduced list, which
gives $\E g=1/m_i$ and $\E h=1/m_k$. Summing over the at most $m_i$ common
keys bounds the expected overlap by $m_i^{-(1-s)/(1+s)}$. The argument
applies to hardmax indicators as well.

On its $r$-dimensional image, $BG$ is a nondegenerate Gaussian vector whose
covariance has largest eigenvalue one. Restricted to radii at least $R$,
the law of the direction $BG/\|BG\|$ has a density with respect to the
uniform probability on the unit sphere that is bounded by
\begin{equation}\label{eq:angular-tail}
 C_B(1+R)^r e^{-R^2/2}.
\end{equation}
This follows by integrating the Gaussian density in polar coordinates,
which bounds the radial integral by $C_B\int_R^\infty z^{r-1}e^{-z^2/2}\,dz$. (For $r=1$ the
uniform probability is that on the two unit directions.) For two
independent uniform directions and $m\ge2$,
\[
 \iint m^{-(1-|u^\top v|)/(1+|u^\top v|)}\,d\sigma(u)d\sigma(v)
          \le C_r(1+\log m)^{-(r-1)/2}.
\]
To see this for $r\ge2$, let $\theta\in[0,\pi/2]$ be the smaller of the
angles between $v$ and $\pm u$; the exponent then equals
$\tan^2(\theta/2)\ge\theta^2/4$, and the density of $\theta$ is at most
$C_r\theta^{r-2}$. For $r=1$ the bound is trivial. Taking $R=0$ therefore
gives $\E\|p\|^2=O(a_n^{-(r-1)})$ for every strength: the diagonal terms
cost $1/n$, the pairs involving one of the first $\sqrt n$ rows of a causal
mask cost $O(n^{-1/2})$, and the removal error is polynomially small on the
remaining pairs.

\paragraph{Below the freezing scale.}
Suppose now that $T_n\to\infty$ and $H_n\to\infty$. Take
$R=(1-\delta)a_n/H_n$ with $0<\delta<1/4$. Discard the rows before position
$n^{1-\xi}$, at cost $O(n^{-\xi})$, where the fixed $\xi>0$ is so small that
$a_n(1-\delta)/a_{m_i}\le1-\delta/2$ on the remaining rows for large $n$.
If either query radius is below $R$, the Cauchy--Schwarz inequality for the
two row vectors and \eqref{eq:diffuse} give a polynomially small
contribution after removal. If both radii exceed $R$, we use that the two
query vectors are independent: applying \eqref{eq:angular-tail} to each and
integrating \eqref{eq:angular-hyper} gives
\begin{equation}\label{eq:meso-upper}
 \E\|p\|^2\le O(n^{-\kappa})+
 C_{B,\delta} a_n^{-(r-1)}
     \exp[-2(1-\delta)^2T_n+O_r(\log(1+T_n))] .
\end{equation}
Here and below, $\kappa>0$ may change with the fixed tolerances. For a
single row, the same split and \eqref{eq:diffuse} bound the expectations of
$n^{-1}\sum_i\rho_i^a$, $a=1/2,1$, by
$O(n^{-\kappa})+\exp[-(1-\delta)^2T_n+O_r(\log(1+T_n))]$. Since
$T_n=o(\log n)$, polynomially small remainders are negligible even after
multiplication by $a_n^{r-1}$. This gives the required upper bounds in
expectation, and Markov's inequality transfers them to probability on the
$T_n$ scale.

\subsection{The matching lower bound from the number of extreme keys}

We now consider late queries, $i\ge n/2$, with
$H_n\|BG_i\|\ge(1+\delta)a_n$, and write
$p_*=\Pp\{H_n\|BG\|\ge(1+\delta)a_n\}$ for their probability. Below the
freezing scale,
\[
 \log p_* = -(1+\delta)^2T_n+O_B(\log(1+T_n)).
\]
The upper bound is \eqref{eq:angular-tail}; for the lower bound it suffices
to keep a single standard normal coordinate along a top singular direction.
When $\liminf H_n/a_n>0$, $p_*$ is bounded below by a positive constant. In
either regime $np_*\to\infty$, so the number of these queries divided by
$np_*$ tends to $1/2$ in probability.

Let $\Pi_B$ be the orthogonal projector onto $\operatorname{im}B$, and
consider the common set
\[
 \mathcal J_n=\{j:\|\Pi_B G_j\|\ge \zeta_K\}.
\]
Since every query direction lies in $\operatorname{im}B$, this set contains
every key whose projection on some query direction exceeds the cutoff
$\zeta_K$. Its expected cardinality is at most $C_{r,K}a_n^{r-1}$: the
projected norm has a chi distribution with $r$ degrees of freedom, whose
tail at $z$ is $O_r(z^{r-2}e^{-z^2/2})$ as $z\to\infty$, and
$ne^{-z_n^2/2}$ is of order $z_n$.

Fix an arbitrarily small failure probability, and apply the
concentrated-row estimate with a sufficiently small $\epsilon$. Conditional
on any qualifying query, its reduced row puts, except with probability
$\epsilon$, at least half of its weight on at most $D$ keys with high
projections, and these keys belong to $\mathcal J_n$. For the original
rows, the expected number of qualifying rows whose removed mass exceeds
$1/4$ is $O(n^{1-\kappa})=o(np_*)$. Markov's inequality for both bad counts,
together with the concentration of the number of qualifying queries, shows
that with the prescribed probability at least $c_0np_*$ original rows each
put at least $1/4$ of their mass on at most $D$ keys of $\mathcal J_n$, for
a numerical constant $c_0>0$. This argument uses only marginal failure
bounds for individual rows, so no independence between rows is needed. By
Markov's inequality, also $|\mathcal J_n|\le C a_n^{r-1}$ with the
prescribed probability. On the intersection of these events, the
Cauchy--Schwarz inequality gives
\begin{equation}\label{eq:shell-lower}
 \frac1n\sum_i\rho_i^a\ge c\,p_*\quad(a=1/2,1),\qquad
 \|p\|^2\ge\frac{(\sum_{j\in\mathcal J_n}p_j)^2}{|\mathcal J_n|}
              \ge c\,a_n^{-(r-1)}p_*^2 .
\end{equation}
The constants may depend on the failure probability and on $\delta$;
taking the failure probability fixed below $1/2$ also gives lower bounds in
expectation. Letting $\delta\downarrow0$ after $n\to\infty$ matches
\eqref{eq:meso-upper}, even when $T_n$ grows more slowly than
$\log\log n$.

Finally, $\E_Z M_n=\sqrt{2/\pi}\,n^{-1}\sum_i\sqrt{\rho_i}$ and
$\E_Z Q_n=n^{-1}\sum_i\rho_i$, while, conditionally on the scores, $P_n$
has the law of $\|p\|^2\chi^2_1$. For $M_n$ and $Q_n$, the conditional upper
and small-ball bounds of Section~\ref{sec:typical} make the ratios to these
conditional means, and their reciprocals, tight uniformly in the weight
matrix. This proves the order estimates in parts~1 and~2 of
Theorem~\ref{thm:growing}.

\paragraph{Concentration of the tokenwise risk for $r\ge2$.}
Write $\bar\rho_n=n^{-1}\sum_i\rho_i$. Conditional Gaussian moments give
\[
 \operatorname{Var}_Z(Q_n)=\frac{2}{n^2}\sum_{i,k}
          \langle w_i,w_k\rangle^2\le 2\|p\|^2 ,
\]
and the expectation of the right-hand side vanishes by the angular upper
bound. It therefore remains to show that $\bar\rho_n\to q_c(B)$ in $L^2$.
For $c=0$ and $c=\infty$ this follows directly from $0\le\bar\rho_n\le1$
and the limit of the first moment. For finite $c>0$ it suffices to consider
two distinct rows whose indices are asymptotic to positive fractions of
$n$, to remove their query keys, and to show that the expected product of
their participations tends to $q_c(B)^2$; after truncation at $\epsilon n$,
the diagonal and early pairs cost $O(1/n+\epsilon)$.

Condition on the two queries. Because $r\ge2$, their directions have inner
product $\varrho<1$ almost surely. The marginal row expectations have the
pointwise limits proved above; in particular, if either limiting row
temperature $c\|BG_i\|$ is at most one, the expected participation of that
row tends to zero, and so does the expected product. Queries with
$1<c\|BG_i\|<1+2\delta$ have probability tending to zero as
$\delta\downarrow0$, because $\|BG_i\|$ has a continuous law, so we may
assume that $c\|BG_i\|\ge1+2\delta$ for both queries. Truncate each
projection list from below at $\zeta_K$, replacing the projections below
the threshold by a null symbol, renormalize the retained weights, and
assign participation zero to an empty list. By the estimate on the mass of
high projections, a fixed sufficiently large $K$ makes the expected
truncation error in either participation arbitrarily small. For any
positive limiting prefix fraction the same proof applies, with $m/n$
bounded below by that fraction.

For a shared key write $\bar p=\Pp(S>\zeta_K)=O_K(1/n)$ and
$\bar p_{12}=\Pp(S>\zeta_K,S'>\zeta_K)$. If $\varrho>-1$, then
$\bar p_{12}\le\Pp(S+S'>2\zeta_K)\le n^{-2/(1+\varrho)+o(1)}$; if
$\varrho=-1$, then $\bar p_{12}=0$. The joint law of the truncated pair is
within total variation distance $O(\bar p_{12}+\bar p^2)$ of the product of
its marginals. On the region where both coordinates exceed the threshold,
the distance is bounded by the sum of these two masses; on each region where
only one coordinate exceeds it, the measure equals the corresponding
marginal tail measure minus a projection of the double-exceedance measure,
which gives the same bound; and the remaining atom is controlled in the same
way. By independence across keys, the two entire lists are within total
variation distance $O(n(\bar p_{12}+\bar p^2))=o(1)$ of independent lists.
Hence the truncated participations of the two rows, which are bounded
functions of the lists, factor in expectation. Letting the truncation error
vanish and integrating over the two independent queries gives $q_c(B)^2$.
This proves the $L^2$ assertion.

In rank one the opposite conclusion holds. Jensen's inequality over the
$n^2$ pairs of rows gives $\operatorname{Var}_Z(Q_n)\ge2\|p\|^4$, and the
law of total variance together with Jensen's inequality again gives
$\operatorname{Var}(Q_n)\ge2(\E P_n)^2$. The lower bound
\eqref{eq:shell-lower} therefore rules out concentration whenever
$\liminf H_n/a_n>0$.

\subsection{The output along score directions}

For $m$ Gaussian keys set
\[
 \Lambda_m(y)=a_m^{-2}\log\!\left(m^{-1}\sum_{j\le m}
                          e^{a_m y^\top X_j}\right).
\]
By Lemma~\ref{lem:laplace}, $\Lambda_m$ converges in probability, uniformly
on every compact set, to $\Lambda(y)=(f(\|y\|)-1)/2$. The gradient of
$\Lambda$ is $\pi(y)$, including at $\|y\|=1$, while the gradient of
$\Lambda_m$ is the softmax-weighted key divided by $a_m$. To show that
these gradients converge uniformly on a compact ball, bound each partial
derivative of the convex function $\Lambda_m$ from below and above by the
backward and forward differences with step $h$ in that coordinate. On the
ball enlarged by $h$, either difference deviates from the corresponding
difference of $\Lambda$ by at most $2\sup|\Lambda_m-\Lambda|/h$, and since
$\pi$ is $1$-Lipschitz, the finite differences of $\Lambda$ deviate from its
partial derivatives by at most $h$. Letting first $m\to\infty$ and then
$h\downarrow0$ gives the claim.

The convergence is in fact uniform over all of $\R^d$. For each unit vector
$u$ and every $t\ge1$, convexity along the ray gives
$u^\top\nabla\Lambda_m(tu)\ge u^\top\nabla\Lambda_m(u)=1-o_{\mathbb P}(1)$,
where the error is uniform in $u$ by the compact case. The norm of the
gradient is at most $K_m=\max_{j\le m}\|X_j\|/a_m$, and $K_m\to1$ in
probability by Gaussian tails. Therefore
$\|\nabla\Lambda_m(tu)-u\|^2\le K_m^2+1-2u^\top\nabla\Lambda_m(tu)$ is
bounded above by an $o_{\mathbb P}(1)$ term, uniformly over these rays.

Now consider a row $i$, with $m=n$ (full attention) or $m=i$ (causal
attention) and $i/n\to s>0$, and evaluate at $y=H_nBX_i/a_m$; the
uniformity allows $X_i$ to be one of the keys. For $d_m=a_m/a_n$, the map
$y\mapsto d_m\pi(y/d_m)$ is the projection onto the ball of radius $d_m$,
so it differs from $\pi(y)$ by at most $|1-d_m|$, uniformly in $y$. Since
$d_m\to1$, the normalized output $O_i/a_n$ differs from $V\pi(c_nBX_i)$ by
a quantity that tends to zero in probability, and by the subsequence
argument used for the row risk this holds uniformly over $i\ge\epsilon n$.
The bound $\|\nabla\Lambda_m\|\le\max_{j\le m}\|X_j\|/a_m$ gives bounded
fourth moments on such prefixes, hence convergence in squared mean. Using
the maximum over all $n$ keys and the scale $a_n$, the first $\epsilon n$
rows contribute $O(\epsilon)$ to the expected normalized average error.
This proves the first assertion of Theorem~\ref{thm:clipping}.

For finite $c$, since $\pi$ is $1$-Lipschitz, replacing $c_n$ by $c$ costs
$o(1)$ in squared mean. Subtracting the normalized population prediction,
using the Cauchy--Schwarz inequality on the difference of squared norms and
applying the $L^1$ law of large numbers to $\|V(\pi(cBX_i)-cBX_i)\|^2$ gives
the second assertion. If $VB\ne0$, a set of Gaussian inputs of positive
measure has both $\|cBX\|>1$ and $VBX\ne0$, so the limit is positive; if
$VB=0$, the two vectors $V\pi(cBX)$ and $cVBX$ are equal. For
$c_n\to\infty$, the bound on the maximum gives
$\E\,n^{-1}\sum_i\|O_i/H_n\|^2=O(c_n^{-2})$, and the Cauchy--Schwarz
inequality and the law of large numbers for $\|VBX_i\|^2$ give the last
limit.

\subsection{Rank-one causal records}

Write $B=uv^\top$ with unit vectors $u,v$. The hardmax winner of a row is
the key with the largest $u^\top G_j$ in its visible prefix if
$v^\top G_i>0$, and the key with the smallest $u^\top G_j$ if
$v^\top G_i<0$; the extrema are almost surely unique. For a row at finite
strength, the conditional squared difference between its output and the
hardmax output of the marked head is at most
$2(1-\max_j w_{ij})^2\le2(1-\rho_i)$. Since the expected participation
tends to $q_\infty=1$, Jensen's inequality shows that the soft and hard
pooled outputs are equivalent in squared mean.

Under full attention, the fraction of queries with positive sign tends to
$1/2$, and the two distinct extreme keys carry independent standard normal
marks conditional on the score coordinates; hence the conditional variance
of the mean tends to $1/2$. For causal attention, order the $n$ keys from
the highest to the lowest projection, and record their positions in $[0,1]$
by dividing the indices by $n$. For any fixed $k$, the positions of the top
$k$ and the bottom $k$ keys form a sample without replacement from a uniform
grid, and therefore converge jointly to $2k$ independent uniform variables.
The marks of these keys are independent standard normal variables,
independent of the positions.

Fix $\epsilon>0$. Except with probability at most $2(1-\epsilon)^k+o(1)$,
the first $\epsilon n$ positions contain both a top-$k$ and a bottom-$k$
key, and then the two extreme indices of every later prefix are chosen from
these two finite lists. Over an interval of positions with the same
positive winner, the number of positive queries divided by $n$ differs from
half the relative length of the interval by $o_{\mathbb P}(1)$, and the same
holds for negative winners. This follows from the maximal inequality for
sums of i.i.d.\ centered signs: the maximum over all prefixes of the
absolute centered sign count, divided by $n$, tends to zero, and this bound
applies to random intervals as well, including the case in which $u$ and
$v$ are parallel. For fixed $k$, the interval lengths are almost surely continuous
functions of the limiting distinct positions, apart from null boundary
ties. Finally, since every individual hardmax output has second moment one,
the contribution of the first $\epsilon n$ rows has $L^2$ norm at most
$\epsilon+1/n$.

Now let $k\to\infty$ at fixed $\epsilon$, and then let $\epsilon\downarrow0$.
In the limiting list of positions of the largest keys, the largest key
visible at time $s$ is the first key in the list whose position is at most
$s$.
Let $R_1$ be the position of the first key in the list; it holds the role
of maximum on an interval of length $R_0-R_1=1-R_1$. In general, let
$R_{j+1}$ be the position of the first later key in the list whose position
lies below $R_j$; this key is the maximum on an interval of length
$R_j-R_{j+1}$. Conditional on the past, $R_{j+1}$ is uniform on $[0,R_j]$,
so the successive ratios are independent uniform variables, and the
interval lengths sum to one almost surely. The list of the smallest keys
gives an independent copy. The limiting pooled output is therefore half the
sum of the independent Gaussian marks on both lists, each weighted by the
length of its interval. The series converges in $L^2$ and has conditional
variance $W$, as stated. Both the finite and the limiting conditional
variances are bounded by one, so by uniform Gaussian moment bounds the
convergence in distribution yields convergence of all fixed moments.

The same interval argument applied to the squared marks gives the full
hardmax limit $Q_n\Rightarrow (Z_+^2+Z_-^2)/2$, an exponential variable with
mean one. In the causal limit, each squared mark has coefficient equal to
half the length of its interval, so the conditional mean of the sum is one
and its conditional variance is $2W$. Since each squared mark has mean one,
truncating the early rows costs at most $\epsilon+1/n$ in $L^1$, and by the
Cauchy--Schwarz inequality and the preceding row coupling, the soft and
hard averages of squares are equivalent in $L^1$. Jensen's inequality gives
$\E Q_n^k\le\E|Z|^{2k}$ for every integer $k\ge1$ and both types of
weights, so under the causal mask the unconditional variance converges to
$2\E W=1/2$.

Finally, $S\stackrel{d}{=}(1-U)^2+U^2 S'$, where $S'$ is an independent copy
of $S$. Taking first and second moments gives $\E S=1/2$ and
$\E S^2=7/24$, hence $\E W=1/4$ and $\operatorname{Var}(W)=1/192>0$. This
proves Theorem~\ref{thm:records}.

As a finite-sample check, suppose that $u\perp v$, so that the query signs
are independent of the entire key list. For $2\le i<k$, two hardmax rows
select the same key with probability $i/(2k)$: their signs must agree, and
the relevant extreme of the first $k$ keys must lie among the first $i$.
For $i=1$ the probability is $1/k$. Summing over all pairs, including the
diagonal, gives the exact causal risk $1/4+3/(4n)+(h_n-1)/n^2$ and the exact
full risk $1/2+1/(2n)$, where $h_n=\sum_{i=1}^n 1/i$.

\subsection{When the value is the key projection}

Set $M_i^+=\max_{j\le i}S_j$, $M_i^-=\max_{j\le i}(-S_j)$,
$\Delta_i=M_i^++M_i^-$ and $\varepsilon_i=\operatorname{sign}(v^\top X_i)$.
The causal hardmax output at position $i$ is
$(M_i^+-M_i^-+\varepsilon_i\Delta_i)/2$. Since the range $\Delta_i$ is
nonnegative and nondecreasing in $i$, summation by parts bounds
$|\sum_{i=1}^n\varepsilon_i\Delta_i|$ by
$2\Delta_n\max_{k\le n}|\sum_{i\le k}\varepsilon_i|$. By Gaussian tails and
the martingale maximal inequality, the two factors have fourth-moment norms
$O(a_n)$ and $O(\sqrt n)$, so the contribution of the sign term to
$a_n\overline O$ vanishes in $L^2$. For full attention the same argument
applies with every $\Delta_i$ replaced by $\Delta_n$.

Recall $z_m=\Phi^{-1}(1-1/m)$. For either sign,
\[
 \sup_{m\ge3}\E|z_m(M_m^\pm-z_m)|^4<\infty .
\]
To see this, write $z=z_m$. The upper tail of the scaled maximum at $x\ge0$
is at most $e^{-x}$, because $m\overline\Phi(z+x/z)\le e^{-x}$. For
$0\le x\le z^2$, shifting the Gaussian integral gives
$m\overline\Phi(z-x/z)\ge e^{x-x^2/(2z^2)}\ge e^{x/2}$, so the lower tail is
at most $\exp(-e^{x/2})$; for $x=z^2+zy$ with $y\ge0$ it is at most
$2^{-m}e^{-my^2/2}$. Integrating these tail bounds proves the moment bound.

For each time interval $I$ of fixed positive length $|I|$ and fixed levels
$x,y$, the probability that no key in $I$ exceeds $z_n+x/z_n$ and none
falls below $-z_n-y/z_n$ tends to $\exp[-|I|(e^{-x}+e^{-y})]$: raise the
probability of neither event for a single key to the number of indices in
the interval. Disjoint intervals involve independent keys, so a finite
partition of time gives the joint finite-dimensional convergence of
$z_n(M_{\lfloor ns\rfloor}^\pm-z_n)$ to the two independent running maxima
$\mathcal M_\pm(s)$ of Theorem~\ref{thm:aligned}.

On $[\epsilon,1]$ the scaled prefix maxima are monotone in time. Their
integral therefore lies between the endpoint Riemann sums, whose difference
is at most the mesh times the increment over the interval; this increment
is tight by the endpoint limits, which gives joint convergence of the
integrals over $[\epsilon,1]$. For $i\ge\sqrt n$ the ratio $z_n/z_i$ is
bounded, and the fourth-moment bound at $i$ gives
$\|z_n(M_i^+-M_i^-)\|_{L^4}=O(1)$, because the two centerings cancel. By
the bound on the maximum, positions $i<\sqrt n$ contribute
$O(n^{-1/2}\log n)$ in $L^4$ to the normalized average, and the remaining
positions before $\epsilon n$ cost $O(\epsilon)$. The limiting processes
have marginal laws $\log s+\mathrm{Gumbel}$, so the integral of their
difference over $(0,\epsilon)$ also costs $O(\epsilon)$ in $L^4$. Together
with $z_n/a_n\to1$, this proves the weak limits for both masks, and uniform
integrability gives convergence of the second moments of the midrange
averages. The omitted sign term vanishes in $L^2$, as shown above.

For $0<s\le t$, write $q=s/t$ and let $C(q)$ denote the covariance of
$\mathcal M_+(s)$ and $\mathcal M_+(t)$, which depends only on $q$. The
covariance equals the integral, over both thresholds, of the joint
distribution function minus the product of the marginal ones (truncate both
variables first and then pass to the limit). With $a=e^{-x}$ and $b=e^{-y}$
the integrand becomes $e^{-sa-tb}(\exp(s\min(a,b))-1)\,da\,db/(ab)$.
Substituting $a=kb$ and integrating over $b$ gives
\[
 C(q)=\int_0^1\frac{\log(1+qk)}{k}\,dk+
 \int_0^1\log\!\left(\frac{q+y}{q+(1-q)y}\right)\frac{dy}{y},
\]
where $y=1/k$ on the part $k\ge1$. In the original $k$-integral the
integrand is increasing in $q$, integrable at $q=1$ and vanishes as
$q\downarrow0$; hence $C(0)=0$. The derivative of the first term is
$\log(1+q)/q$, and that of the second is the integral of
$q^{-1}[(q+(1-q)y)^{-1}-(q+y)^{-1}]$; summing gives
$C'(q)=-\log q/(1-q)$ for $0<q<1$. Integrating by parts, we obtain
\[
 \operatorname{Var}\!\left(\int_0^1\mathcal M_+(s)\,ds\right)
 =2\int_0^1\int_0^t C(s/t)\,ds\,dt
 =\int_0^1 C(q)\,dq=\int_0^1-\log q\,dq=1.
\]
At a single time, $C(1)=\sum_{k\ge1}k^{-2}=\pi^2/6$. Since the two signs
are independent, the limiting variances on the $a_n$ scale are $1/2$ for
the causal mask and $\pi^2/12$ for the full mask, which gives the constants
in Theorem~\ref{thm:aligned}.

Finally, for a softmax with inverse temperature $h>0$ on at most $n$ scalar
scores, the weighted score falls short of the maximum by at most
$\log n/h$. This follows by writing the log-partition function as $h$ times
the weighted score plus the entropy of the weights, which is at most
$\log n$; reversing signs gives the same bound at the minimum. Set
$\tau=(\log n)^{-2}$, $T_i=|v^\top X_i|$ and $N_\tau=\#\{i:T_i\le\tau\}$.
For either mask, the scaled difference between the soft and the hard pooled
outputs is bounded in absolute value by
\[
 a_n \Delta_n N_\tau/n+
 \frac{a_n\log n}{H_n}\frac1n\sum_i T_i^{-1}\ind\{T_i>\tau\}.
\]
The binomial count satisfies $\|N_\tau/n\|_{L^4}=O(\tau)$, because
$n\tau\to\infty$ and the half-normal density is bounded at zero, so the
first term tends to zero in $L^2$ by the Cauchy--Schwarz inequality.
Integrating its first two moments and using independence across $i$ shows
that the average in the second term has $L^2$ norm
$O(1+|\log\tau|+(n\tau)^{-1/2})$. The growth condition
$H_n/(a_n\log n\log\log n)\to\infty$ therefore makes the second term vanish
in $L^2$, which proves the softmax extension.

\subsection{Bounds uniform over the singular spectrum}\label{sec:spectrum-proof}

Write $B=B_n$, drop the subscript $n$ from its singular values, and put
$N(a)=\prod_{j=2}^d(1+a\sigma_j)$, so that $N_n=N(a_n)$. The estimates allow
zero singular values, and for $d=1$ the product is empty.

\paragraph{An upper bound for the mass of an angular cap.}
Let $G$ be standard Gaussian and $U=BG/\|BG\|$. For a unit vector $v$ write
$\theta(U,v)=\arccos|U^\top v|$ for the acute angle between $U$ and the
line through $v$. We claim that, for $t>0$,
\begin{equation}\label{eq:spectral-cap}
 \sup_v\Pp\{\theta(U,v)\le t\}\le C_d/N(1/t).
\end{equation}
Let $\Pi$ be the orthogonal projection onto $v^\perp$. On the event
$\{\theta(U,v)\le t\}$ we have $\|\Pi BG\|\le t\|BG\|$, and hence
$\|\Pi B(G/\|G\|)\|\le t$, because $\|B\|_{\rm op}=1$. By rank-one
interlacing, applied to $(\Pi B)^\top\Pi B=B^\top B-B^\top vv^\top B$, the
$j$th singular value of $\Pi B$ is at least $\sigma_{j+1}$. In the right
singular basis of $\Pi B$, the vector $G/\|G\|$ is uniform on the sphere.
Let $k$ be the number of indices $j$ with $\sigma_{j+1}>2\sqrt d\,t$; on the
event above, the first $k$ coordinates $y_j$ of $G/\|G\|$ in this basis
satisfy $|y_j|\le t/\sigma_{j+1}$. This box lies in the $k$-dimensional ball
of radius $1/2$, where the joint density of the first $k$ coordinates, which
is proportional to $(1-\|y\|^2)^{(d-k-2)/2}$ for $1\le k\le d-1$, is bounded
by a constant depending only on $d$. Integrating gives the bound
$C_d\prod_{j=1}^k t/\sigma_{j+1}$. The omitted factors $1+\sigma_{j+1}/t$,
$j>k$, are at most $1+2\sqrt d$, and for $k=0$ the trivial bound one
suffices. This proves \eqref{eq:spectral-cap}.

For two independent directions $U,U'$, write $\theta=\theta(U,U')$, let
$m$ be the shorter of the two reduced prefix lengths, and put
$L=\log m$. By the overlap bound $m^{-(1-s)/(1+s)}$ of
Section~\ref{sec:angle}, with $s=\cos\theta$, and by
$(1-s)/(1+s)=\tan^2(\theta/2)\ge\theta^2/4$, the conditional expectation of
the overlap is at most $e^{-\theta^2L/4}$. Since
$e^{-\theta^2L/4}=\int_0^\infty 2xe^{-x^2}\,\ind\{\theta\le2x/\sqrt L\}\,dx$,
applying \eqref{eq:spectral-cap} with $t=2x/\sqrt L$ gives the bound
$C_d/N(\sqrt L)$, because $N(a)/N(a/y)\le\max(1,y)^{d-1}$. Once the first
$\sqrt n$ positions are dropped, $\sqrt{\log m}$ is comparable to
$a_n$, and the distinguished-key error from Section~\ref{sec:growing-proof}
is polynomially small, uniformly in $B$ and $H_n$. The diagonal and early
pairs also have polynomially small total weight. Since
$N(a_n)\le(1+a_n)^{d-1}$, these polynomially small errors are negligible
relative to $1/N(a_n)$, which proves the upper bound for every strength.

Below the freezing scale we restrict both query radii to
$\|BG\|\ge R=(1-\delta)a_n/H_n$, with a fixed $0<\delta<1/4$. On this event
$\|G\|\ge R$. For $\eta=(1+R^2)^{-1}$, the standard Gaussian density on
$\{\|G\|\ge R\}$ is bounded by $\eta^{-d/2}e^{-(1-\eta)R^2/2}$ times the
density of $\mathcal N(0,I_d/\eta)$, and a scalar dilation does not change
the direction of $BG$. Since
$\eta^{-d/2}e^{-(1-\eta)R^2/2}=(1+R^2)^{d/2}e^{-R^2/2}e^{R^2/(2(1+R^2))}
\le e^{1/2}(1+R)^de^{-R^2/2}$, restricting to the radial event multiplies
the measure of any set of directions by at most $C_d(1+R)^d e^{-R^2/2}$.
Applying this to both queries preserves the cap factor. As in
\eqref{eq:meso-upper}, we discard the first $n^{1-\xi}$ positions for a
small fixed $\xi>0$; if either radius is below $R$, the reduced row of that
query satisfies \eqref{eq:diffuse}, which makes the overlap polynomially
small. Hence
\[
 \E P_n\le O(n^{-\kappa})+
 \frac{\exp[-2(1-\delta)^2T_n+O_d(\log(1+T_n))]}{N(a_n)}.
\]

\paragraph{A small common set of keys.}
In the right singular basis, select the late queries ($i\ge n/2$) with
$b\le G_{i,1}\le b+1/(1+b)$ and $|G_{i,j}|\le1$ for $j\ge2$, where
$b=\max(1,(1+\delta)a_n/H_n)$ for softmax and $b=1$ for hardmax.
Integrating the normal density over these intervals shows that their
probability $p_b$ satisfies $\log p_b=-b^2/2+O_d(\log(1+b))$. In the left
singular basis, their directions belong to
\[
 \mathcal U=\{(1,y)/\|(1,y)\|:\ |y_j|\le\sigma_j,\ 2\le j\le d\},
\]
and for finite $H_n$ they have score norm at least $(1+\delta)a_n$. Let
$z=z_n=\Phi^{-1}(1-1/n)$. A rectangular grid of mesh $1/z$ in the
coordinates $y$ covers $\mathcal U$ by $O_d(N(z))$ caps of angular radius
$L_d/z$, because the normalization map is Lipschitz on these coordinates,
where $\|(1,y)\|\ge1$.

Let $\mathcal J$ be the set of keys whose projection along some direction
in $\mathcal U$ is at least $z-K/z$. A single cap of the net contributes at
most $C_{K,d}$ keys in expectation. To see this, decompose a fresh key along
the center of the cap as $(S,X_\perp)$, with independent Gaussian
components. For large $n$, its projection on any direction in the cap is at
most $S_++L_d\|X_\perp\|/z$. Put $K'=K+L_d\|X_\perp\|$. If $K'\le z^2/2$,
exceeding the cutoff requires $S\ge z-K'/z$, which by Mills' inequality has
probability at most $Ce^{K'}\,\overline\Phi(z)$. Moreover
$\E e^{K'}<\infty$ and $\Pp(K'>z^2/2)=O(e^{-c_d z^4})$. Since
$n\overline\Phi(z)=1$ and $z/a_n\to1$, summing over the grid gives
$\E|\mathcal J|\le C_{K,d}N(a_n)$.

Since $np_b\to\infty$, there are $(1/2+o_{\mathbb P}(1))np_b$ qualifying
late queries. Conditional on such a query, the concentrated-row estimate for
the reduced prefix shows that, with probability arbitrarily close to one
for $K$ large, at least half of the weight of the row lies on
$\mathcal J$; hardmax satisfies the same estimate because it selects the
maximal projection. The expected number of qualifying rows in which the
distinguished keys carry mass above $1/4$ is $O(n^{1-\kappa})=o(np_b)$.
Markov's inequality for these two bad counts and for $|\mathcal J|$ gives,
with any prescribed probability close to one,
\[
 \sum_{j\in\mathcal J}p_j\ge c p_b,\qquad
 \sum_jp_j^2\ge \frac{(\sum_{j\in\mathcal J}p_j)^2}{|\mathcal J|}
                   \ge c'\,p_b^2/N(a_n).
\]
This argument requires no independence between the events for different
rows, or between these events and $|\mathcal J|$. For $H_n\ge ha_n$, $p_b$
is bounded below in terms of $h$, $\delta$ and $d$; below the freezing
scale, $\log p_b=-(1+\delta)^2T_n+O_d(\log(1+T_n))$. A fixed success
probability gives the lower bounds in expectation. In the upper bound, the
polynomially small remainder is negligible because $T_n=o(\log n)$ and
$N(a_n)\le(1+a_n)^{d-1}$, and Markov's inequality transfers that bound to
probability at the cost of a factor $e^{\eta T_n}$, for any $\eta>0$. For
$P_n$ itself we use its conditional law $\|p\|^2\chi^2_1$, which preserves
tightness and the logarithmic rate. Letting $\delta\downarrow0$ after
$n\to\infty$ proves the bounds on the pooled risk.

\paragraph{Concentration of the average squared row output.}
With Gaussian marks, the arguments of Section~\ref{sec:growing-proof},
based on conditional variances and Jensen's inequality, give
\[
 2(\E P_n)^2\le\operatorname{Var}(Q_n),\qquad
 \E\operatorname{Var}_Z(Q_n)\le2\E P_n .
\]
The first inequality and the lower bound on $\E P_n$ prove necessity: along any subsequence on which
$a_n\sigma_{2,n}$ stays bounded, $N_n$ stays bounded and
$\operatorname{Var}(Q_n)$ is bounded away from zero. For sufficiency,
suppose that $a_n\sigma_{2,n}\to\infty$. Then $N(a_n)\to\infty$, so by the
second inequality it suffices to show that the variance of
$\bar\rho_n=n^{-1}\sum_i\rho(w_i)$ tends to zero. For exact hardmax
$\bar\rho_n=1$. For softmax, pass to a subsequence along which all singular
values converge, to $\sigma_j^\circ$ say, and $H_n/a_n\to c\in[h,\infty]$.
Couple each query to a single standard Gaussian vector $G$ in right singular
coordinates. At any positive prefix fraction the reduced row then has
$t/a_m\to cR$ with $R=\|\operatorname{diag}(\sigma^\circ)G\|>0$ almost
surely, and the single-row limit gives $\E\bar\rho_n\to\E(1-1/(cR))_+$,
interpreted as one for $c=\infty$. In that case $0\le\bar\rho_n\le1$ already
forces the variance to vanish.

For finite $c$, consider distinct rows with positive limiting prefix
fractions, and compare the product of their participations with the same
product computed from independent reduced key lists, conditional on the two
queries. If either limiting value $cR$ is at most one, the expected
participation of that row tends to zero. Since $R$ has a continuous law,
the probability of a query with $1<cR<1+2\delta$ tends to zero with
$\delta$. For the remaining supercritical rows, the concentrated-row
estimate shows that truncating the projections below $\zeta_K=z_n-K/z_n$
changes the expected product arbitrarily little, for fixed $\delta$ and
sufficiently large $K$. These statements hold for independent lists as
well. For a common key let $\bar p=\overline\Phi(\zeta_K)$, and let
$\bar p_{12}$ be the probability of a double exceedance given the angle
$\theta$ between the two query directions. The Gaussian bound for the sum
of the two projections and Mills' inequality give
\[
 n\bar p_{12}\le n\,\overline\Phi\!\left(\zeta_K\sqrt{2/(1+|\cos\theta|)}\right)
       \le C_K e^{-c_0 a_n^2\theta^2}.
\]
Indeed, $n\overline\Phi(z_n)=1$, $z_n^2-\zeta_K^2=O_K(1)$, and
$(1-|\cos\theta|)/(1+|\cos\theta|)\ge\theta^2/4$ for the acute angle.
Integrating the cap bound as above gives
$\E[n\bar p_{12}]\le C_{K,d}/N(a_n)$. Conditional on the queries, the joint
truncated lists differ from independent lists by at most
$Cn(\bar p^2+\bar p_{12})$ in total variation, as in the fixed-rank
argument, and the expectation of this bound tends to zero. Letting the
truncation errors vanish therefore factors the product in expectation,
because the queries are independent. Averaging over pairs of rows, among
which early and coincident pairs have vanishing weight, gives
$\operatorname{Var}(\bar\rho_n)\to0$. Since every subsequence has a further
subsequence of the above form, this proves the variance assertion of
Theorem~\ref{thm:spectrum}.

\subsection{The planar transition for two nested prefixes}\label{sec:prefix-proof}

We first fix the two query directions. After the removal of their own
keys, the shorter and the longer list have lengths $m$ and $k$ with
$m/n\to\alpha$, $k/n\to\beta$ and $0<\alpha\le\beta\le1$; the two lists share
all $m$ keys of the shorter one. Each shared pair of projections $(S,S')$
is bivariate standard normal with correlation $\varrho_n$. Suppose that
$\varrho_n\to1$ and $z_n\sqrt{1-\varrho_n^2}\to s<\infty$. Write $z=z_n$ and
$u_n(x)=z+x/z$. Given $S=u_n(x)$, the rescaled second projection
$z(S'-z)$ converges in law to $x+W$, where $W=sZ-s^2/2$ and
$Z\sim\mathcal N(0,1)$; this uses $z^2(1-\varrho_n)\to s^2/2$.

Let $F_n^{(2)}$ be the joint distribution function of this Gaussian pair.
For $x_n\to x$ and $y_n\to y$, the probability of the union of the two tails
satisfies
\begin{equation}\label{eq:union-limit}
 n[1-F_n^{(2)}(u_n(x_n),u_n(y_n))]
   \longrightarrow e^{-x}L_s(y-x),\qquad
 L_s(w)=\E\max(1,e^{W'-w}),
\end{equation}
where $W'$ has the same law as $W$.
Indeed, $n\phi(u_n(t))/z\to e^{-t}$, with the bound $Ce^{-t}$ on every
half-line that is bounded below, so integrating the conditional exceedance
probability over $t\ge x_n$ gives the limit of the intersection,
$\int_x^\infty e^{-t}\Pp(t+W>y)\,dt$. The two marginal tails have limits
$e^{-x}$ and $e^{-y}$. Since $\E e^W=1$, adding the marginals and
subtracting the intersection gives
$\int e^{-t}\Pp(t>x\text{ or }t+W>y)\,dt=e^{-x}L_s(y-x)$. The argument
includes $s=0$, where one ignores the single boundary value of $t$. This is
the bivariate Gaussian extreme-value limit of H\"usler and
Reiss~\cite{husler}.

Condition on the winning projection $M$ in the shorter list and on the
second projection $Y=\varrho_nM+\sqrt{1-\varrho_n^2}\,Z$ of the same key.
The remaining $m-1$ shared keys have first projections conditioned to lie
below $M$, so the probability that the same key also wins the longer list
is exactly
\[
 \E\left[\left(\frac{F_n^{(2)}(M,Y)}{\Phi(M)}\right)^{m-1}
                         \Phi(Y)^{k-m}\right].
\]
Couple $M=\Phi^{-1}(\omega^{1/m})$ to a single uniform variable $\omega$
independent of $Z$. Then $z(M-z)\to\Gamma$ almost surely, where
$e^{-\Gamma}$ is exponential with rate $\alpha$, and $z(Y-z)\to\Gamma+W$.
By \eqref{eq:union-limit}, the bounded integrand converges to
$\exp[-e^{-\Gamma}\{\alpha(L_s(W)-1)+(\beta-\alpha)e^{-W}\}]$, where $L_s$ is
evaluated at the realized value $W$. Integrating over $\Gamma$ gives
$\E\{1/[L_s(W)+(q^{-1}-1)e^{-W}]\}$ with $q=\alpha/\beta$, and Gaussian
tilting gives $L_s(W)=\Phi(Z)+e^{-W}\Phi(s-Z)$, including the case $s=0$ by
continuity. This proves \eqref{eq:prefix-concurrence}. If instead
$\varrho_n\to-1$, both maxima exceed $z-K/z$ with probability tending to one
as $K\to\infty$ after $n\to\infty$, while for each fixed $K$,
$n\Pp\{S+S'\ge2(z-K/z)\}\to0$; hence the probability of a common winner
vanishes.

For a planar query $(G_1,G_2)$, the sign of the leading coordinate is
independent of the slope $G_2/G_1$, which is standard Cauchy. Conditional
on equal signs, the angle between two query directions is the difference
of the values of $\arctan(\varepsilon_nG_2/G_1)$ for the two slopes. Thus
$z_n\sqrt{1-\varrho_n^2}\to\lambda|D|$ almost surely, where $D$, the
difference of the two slopes, is Cauchy with scale two; opposite signs give
$\varrho_n\to-1$. We now average the preceding bounded probabilities over
the queries and over two independent uniform row indices. By the earlier
bound on any positive prefix fraction, removing the distinguished keys
costs $o(1)$, and the coincident and early pairs of rows have vanishing
total contribution once the cutoff on the prefix fraction is sent to zero.
Full attention has $q=1$. For causal attention, the ratio of the smaller to
the larger of two independent uniform times is uniform on $(0,1)$. The
identity $\E P_n=\Pp\{\text{two sampled rows select the same key}\}$ now
gives both limits.

For softmax with $H_n/a_n\to\infty$, note that $\|B_nG_i\|\ge|G_{i,1}|$.
For a positive prefix fraction, the reduced row therefore has expected
participation tending to one, by \eqref{eq:stein-growing} and dominated
convergence, and removing the distinguished keys preserves this limit.
Conditional on the scores, the squared difference between the output of
the row and its hardmax output is at most $2(1-\rho(w_i))$, so Jensen's
inequality and the uniform Gaussian second moments transfer the limits of
the expected pooled risks. Finally, $K_0(q)=q$, $0<K_s(q)\le1$, and
$K_s(q)<K_s(1)$ for $q<1$; bounded convergence gives continuity. Applying
Theorem~\ref{thm:spectrum} with $\varepsilon_n=\lambda/a_n$ for all
sufficiently large $n$ gives the two orders for large $\lambda$, with
constants independent of $\lambda$.

As a numerical check of the unequal-prefix calculation, we simulated 32,768
independent limiting Poisson fields at $s=1.5$. For $q=0.4$ and $q=1$ the
observed frequencies of a common winner were $0.2064$ and $0.3725$ (Monte
Carlo standard errors $0.0022$ and $0.0027$), while
\eqref{eq:prefix-concurrence} gives $0.2076$ and $0.3776$. The sampler keeps
all points for which either coordinate exceeds $-\log48$; the probability
of missing a prefix maximum above this cutoff is at most
$e^{-48q}+e^{-48}$.

\end{document}